\def\year{2019}\relax
\pdfoutput=1
\documentclass[letterpaper]{article} 
\usepackage{aaai19}  
\usepackage{times}  
\usepackage{helvet} 
\usepackage{courier}  
\usepackage[hyphens]{url}  
\usepackage{graphicx} 
\usepackage{graphicx}  
\usepackage{amsmath}
\usepackage{amsfonts}
\usepackage{nidanfloat}
\usepackage{lscape}
\usepackage{subfigure}
\usepackage[linesnumbered,ruled]{algorithm2e}
\usepackage{algorithmic}
\global\long\def\T#1{#1^{\top}}
\newcommand{\vect}[1]{\mbox{\boldmath $#1$}}
\makeatletter
\newcommand{\figcaption}[1]{\def\@captype{figure}\caption{#1}}
\newcommand{\tblcaption}[1]{\def\@captype{table}\caption{#1}}
\makeatother

\title{Refining Coarse-grained Spatial Data Using Auxiliary Spatial Data Sets \\
with Various Granularities}
\author{
Yusuke Tanaka,\textsuperscript{1,3}
Tomoharu Iwata,\textsuperscript{2}
Toshiyuki Tanaka,\textsuperscript{3}
Takeshi Kurashima,\textsuperscript{1}\\
{\bf \Large
Maya Okawa,\textsuperscript{1}
Hiroyuki Toda\textsuperscript{1}}\\
\textsuperscript{1}{NTT Service Evolution Laboratories, NTT Corporation}\\
\textsuperscript{2}{NTT Communication Science Laboratories, NTT Corporation}\\
\textsuperscript{3}{Graduate School of Informatics, Kyoto University}\\
\{tanaka.y, iwata.tomoharu, kurashima.takeshi, okawa.maya, toda.hiroyuki\}@lab.ntt.co.jp,
tt@i.kyoto-u.ac.jp
}
\begin{document}

\nocopyright
\maketitle

\begin{abstract}
We propose a probabilistic model 
for refining coarse-grained spatial data 
by utilizing auxiliary spatial data sets.
Existing methods require that 
the spatial granularities of the auxiliary data sets 
are the same as the desired granularity of target data.
The proposed model can effectively make use of auxiliary data sets 
with various granularities by hierarchically incorporating Gaussian processes.
With the proposed model, 
a distribution for each auxiliary data set
on the continuous space is modeled using a Gaussian process,
where the representation of uncertainty considers the levels of granularity. 
The fine-grained target data are modeled by 
another Gaussian process that considers
both the spatial correlation 
and the auxiliary data sets with their uncertainty.
We integrate the Gaussian process with 
a spatial aggregation process 
that transforms the fine-grained target data 
into the coarse-grained target data,
by which we can infer the fine-grained target Gaussian process
from the coarse-grained data.
Our model is designed such that the 
inference of model parameters based on the exact marginal likelihood is possible,
in which the variables of fine-grained target and auxiliary data are analytically 
integrated out.
Our experiments on real-world spatial data sets demonstrate 
the effectiveness of the proposed model.
\end{abstract}

\section{Introduction}
\label{sec:introduction}
Many cities around the world are now collecting 
large amounts of spatial data 
from a wide range of sources. 
Governments and other organizations are releasing 
data on items such as poverty rate, 
air pollution, traffic flow, energy consumption and 
crime~\cite{shadbolt:linked,goldstein:beyond,barlacchi:multi}. 
Analyzing such spatial data is 
of critical importance 
in improving the life quality of citizens 
in many fields such as 
socio-economics~\cite{rupasinghaa:social,Smith:poverty}, 
public health~\cite{jerrett:spatial},
public security~\cite{bogomolov:once,wang:crime} 
and urban planning~\cite{yuan:discovering}.
For example, knowing the spatial distribution of poverty 
enables us to optimize allocation of resources 
for remedial action. 
Likewise, 
the spatial distribution of air pollution is useful 
in creating policies 
that can control air quality 
and thus protect human health. 

\begin{figure}
 \vspace{-8pt}
 \begin{tabular}{c}
  \hspace{-5pt}
  \begin{minipage}{\linewidth}
   \begin{center}
   \subfigure[Community]
   {\includegraphics[width=33mm]{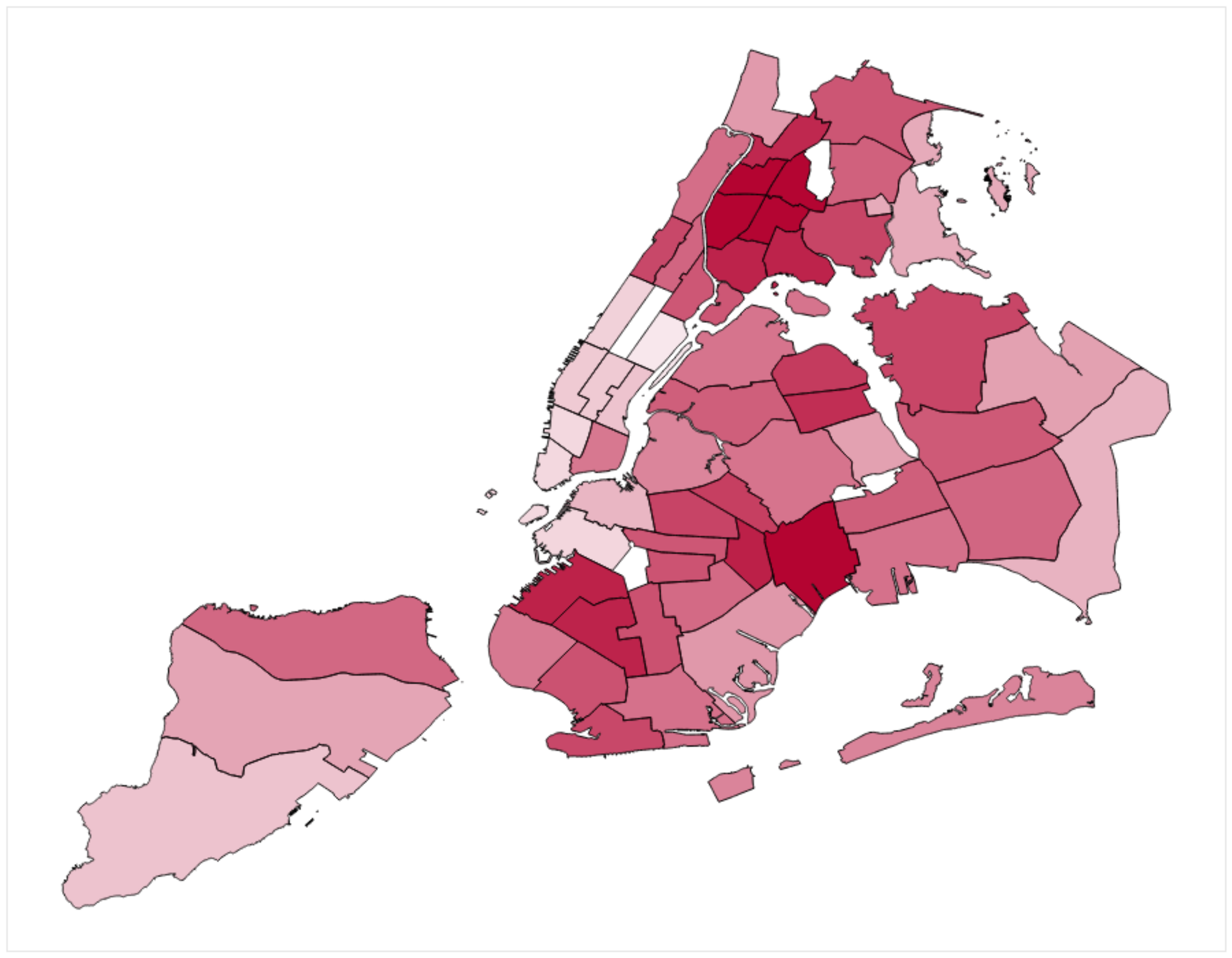}\label{fig:intro_fig_comm}}
   \hspace{5pt}
   \subfigure[Borough]
   {\includegraphics[width=33mm]{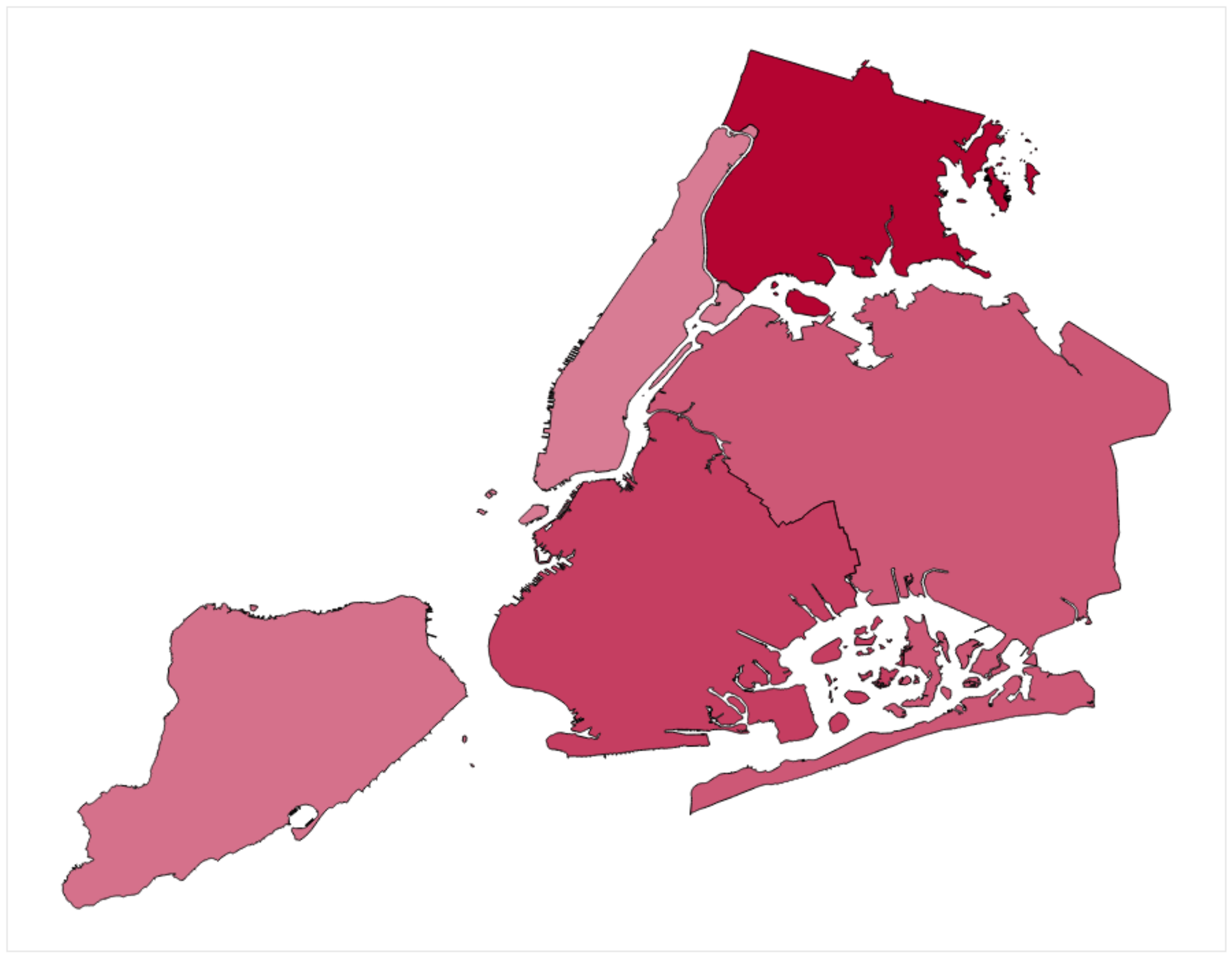}\label{fig:intro_fig_boro}}
   \end{center}
  \end{minipage} 
 \end{tabular}
 \vspace{-17pt}
 \caption{The distribution of poverty rates at different spatial granularities.}
 \label{fig:intro_fig}
 \vspace{-15pt}
\end{figure}
Naturally, information at fine spatial granularity is preferred 
because it allows us to identify 
key regions that require intervention 
to improve city environments efficiently.
As an example, 
Figures~\ref{fig:intro_fig_comm} and \ref{fig:intro_fig_boro}
visualize 
the distributions of poverty rates 
in New York City
by community district and by borough, respectively;
darker hues represent poorer regions.
Clearly, to better understand socio-economic problems, 
Figure~\ref{fig:intro_fig_comm} is better than Figure~\ref{fig:intro_fig_boro}.
In practice, however, such information is often aggregated 
into coarse granularities as in Figure~\ref{fig:intro_fig_boro}.
It is usually thought to be too time-consuming and costly 
to conduct a \textit{census} over the whole population of a city,
and a \textit{sample survey} is conducted instead.
Accordingly, 
the number of samples associated with 
each fine-grained region 
may not be large enough 
to provide a statistically significant estimate
of the value associated to this region;
the typical response is to aggregate samples over larger regions~\cite{Smith:poverty}. 

With the recent increase in data availability, 
utilizing auxiliary spatial data sets on the same region
is an effective way of 
refining coarse-grained target data~\cite{bogomolov:once,park:spatial,smith:beyond,Smith:poverty,wotling:regionalization}. 
In these works, 
the regression models are used for estimating the relationships 
between target data (e.g., poverty rate) 
and auxiliary data sets (e.g., unemployment rate). 
These existing methods, however, require that
the spatial granularities of all the auxiliary data sets 
are the same as the desired granularity of target data. 
This requirement prevents us from making full use of
the auxiliary data sets with various granularities. 
The auxiliary data sets are 
actually associated with various geographical partitions. 
For example, New York City has released 
various spatial data sets portioned into 
boroughs, community districts, zip code, police precincts and so on.

We propose a probabilistic model 
for refining coarse-grained target data
via the effective use of auxiliary data sets 
with various granularities.
An important characteristic is 
discerning the usefulness of each auxiliary data set 
which depends on not only the strength of relationship with the target data 
but also the level of spatial granularity. 
For example, consider the case of two auxiliary data sets
that have the same strength of relationship with the target data, 
but different granularities.
In that case, 
the finer-grained one is seen as more helpful
for refining the coarse-grained target data. 

With the proposed model, 
the fine-grained target data are assumed to follow 
a Gaussian process (GP)~\cite{carl:gaussian} 
whose mean function 
is modeled by a linear regression 
of the auxiliary data sets. 
This GP-based modeling allows us to 
consider the spatial correlation 
in the target data 
and the auxiliary data sets simultaneously.
Since the target data are observed 
not at fine granularity but at coarse granularity, 
we model a spatial aggregation process 
to transform the fine-grained target data 
into the coarse-grained target data. 
Furthermore, 
to handle auxiliary data sets with various granularities, 
we apply GP regression to each auxiliary data set
to derive a predictive distribution
defined on the continuous space;
this conceptually corresponds to spatial interpolation. 
A key idea is that 
it hierarchically incorporates 
the predictive distributions into the model; 
that is, it does not use point estimates. 
This enables us to consider 
uncertainty in the prediction of 
auxiliary data sets. 
The uncertainty is governed 
by several factors, 
one of which is sample density, 
i.e., spatial granularity of the auxiliary data; 
the finer the granularity is, the lower the uncertainty is. 
Incorporating the uncertainty leads to 
effectively learning 
the usefulness of the auxiliary data
with consideration of 
the levels of spatial granularity; 
this allows our model to accurately refine the coarse-grained target data.
We predict the fine-grained target data via a Bayesian inference procedure.
The proposed model is designed such that the
estimation of model parameters based on the exact marginal likelihood is possible:
By analytically integrating out the variables of 
fine-grained target and auxiliary data,
we can estimate the parameters without explicitly obtaining these variables. 
We construct the predictive distribution of the fine-grained target data
by using the estimated parameters.

\section{Related Work}
\label{sec:related}
The problem of refining coarse-grained spatial data 
has been studied in various fields 
such as socio-economics~\cite{smith:beyond,Smith:poverty}, 
agricultural economics~\cite{howitt:spatial,xavier:disaggregating},
epidemiology~\cite{sturrock:fine}
, meteorology~\cite{wilby:guidelines,zorita:analog} and 
geographical information system (GIS)~\cite{boucher:super,goovaerts:combining}.
This problem is also called {\em statistical downscaling}, 
{\em spatial disaggregation}, and {\em areal interpolation}.
The previous works can be categorized into 
two cases in terms of target data availability. 

In the first case, 
in which a large amount of coarse- and fine-grained target data 
are available, 
we can predict 
the fine-grained target data
by using a mapping function 
from coarse- to fine-grained data. 
The mapping function can be learnt 
by using various machine learning methods 
including linear regression models~\cite{hessami:automated}, 
neural networks~\cite{cannon:quantile,misra:statistical} 
and support vector machines~\cite{ghosh:SVM}. 
Recently, 
super-resolution techniques 
based on deep neural networks have been applied 
for refining coarse-grained spatial data~\cite{vandal:deepsd,vandal:generating}. 
The super-resolution techniques aim to 
learn a mapping function 
from low- to high-resolution images~\cite{dong:learning}. 
The method by~\cite{vandal:deepsd} is based on 
the analogy between gridded spatial data and images; 
values at grid cells are regarded as values at pixels. 
The large amount of fine-grained data needed for training are, 
however, not available in many cases (e.g., poverty survey), 
and often only coarse-grained data are available. 
These methods are not applicable in such situations. 

In the second case, 
in which only coarse-grained target data 
are available, 
many regression-based methods have been proposed 
that use auxiliary spatial data sets to refine coarse-grained 
target data~\cite{flaxman:who,Smith:poverty,wang:crime,zheng:U-Air,zheng:forecasting}. 
Regression models (linear and non-linear) are used 
for estimating the relationships 
between target data and auxiliary data sets. 
A few methods can construct the regression models 
under the spatial aggregation constraints~\cite{murakami:new,park:spatial}. 
The constraints state that 
a value associated 
with a coarse-grained region
is a linear average of their constituent values 
in a fine-grained partition. 
In order to satisfy 
the spatial aggregation constraints, 
the regression residuals 
at the coarse-grained regions
are allocated 
to the fine-grained regions
by using the spatial interpolation method, 
i.e., kriging~\cite{stein:interpolation}. 
These methods, however, assume that 
the auxiliary data sets have spatial granularities
equivalent to that of fine-grained target data to be estimated.
This assumption makes it difficult
to utilize multiple auxiliary data sets 
with various granularities. 

Several regression methods have been developed 
for estimating relationships 
between multi-scale spatial data 
sets~\cite{miller:impact,diodato:multiscale,xu:multi,xu:muscat}.
These methods predict the target data 
with the same granularity as that of the training data
by utilizing multi-scale auxiliary data sets. 
They do not, however, consider 
the spatial aggregation constraint,
which is a critical factor in 
refining the coarse-scale target data.

There have been several hierarchical Bayesian models 
to predict fine-grained target data 
using fine-grained auxiliary data sets.
Although they introduce 
a fully Bayesian inference~\cite{taylor:continuous,wilson:pointless,keil:downscaling}
or a variational inference~\cite{Law:variational}
for model parameters,
the uncertainty in the prediction of auxiliary data sets is ignored:
They cannot discern the usefulness of each auxiliary data set 
considering their levels of spatial granularity.

Different from prior works, 
the proposed model can effectively make use of auxiliary data sets 
with various granularities by hierarchically incorporating Gaussian processes.
This hierarchical modeling allows us to effectively learn
the usefulness of each auxiliary data set
considering the levels of spatial granularity. 
Our model also considers 
the spatial aggregation constraints 
by integrating the Gaussian processes with 
a spatial aggregation process to transform the fine-grained target data
into the coarse-grained target data.

\section{Problem Formulation}
\label{sec:problem}
\begin{table}
 \small
 \caption{Notation.}
 \vspace{-15pt}
 \begin{center}
  \begin{tabular}{@{\hspace{3.0pt}} l @{\hspace{8.0pt}} l } \hline
   Symbol&Description  \\ \hline
   ${\mathcal S}$ & set of indices of auxiliary spatial data sets \\
   $s$ & index of auxiliary spatial data set, $s \in {\mathcal S}$ \\
   ${\mathcal X}$ & total region of a city \\
   ${\vect x}$ & location point represented by \\
   & latitude and longitude coordinates, ${\vect x} \in {\mathcal X}$ \\
   ${\mathcal P}^{\rm coar}$ & coarse-grained partition of ${\mathcal X}$ of target data \\
   $i$ & region in the coarse-grained partition \\
   &of target data, $i \in {\mathcal P}^{\rm coar}$ \\
   ${\mathcal P}^{\rm fine}$ & fine-grained partition of ${\mathcal X}$ of target data \\
   $j$ & region in the fine-grained partition \\
   &of target data, $j \in {\mathcal P}^{\rm fine}$\\
   ${\mathcal P_s}$ & partition of ${\mathcal X}$ of $s$th auxiliary data set \\
   $p$ & region in the partition of $s$th auxiliary data set, $p \in {\mathcal P_s}$\\
   $a_i$ & value associated with region $i$ in coarse-grained\\
   &target data, $a_i \in {\mathbb R}$ \\
   $z_j$ & value associated with region $j$ in fine-grained\\
   &target data, $z_j \in {\mathbb R}$ \\
   $y_{s,p}$ & value associated with region $p$ in $s$th auxiliary \\
   &data set, $y_{s,p} \in {\mathbb R}$ \\
   \hline
  \end{tabular}
  \label{notation}
 \end{center}
\vspace{-15pt}
\end{table}
In this section, we describe the spatial data this study focuses on, 
and define our problem of refining coarse-grained spatial data 
by using, for the same region, auxiliary spatial data sets 
with various granularities.
Assume that we have a target spatial data set with coarse granularity, 
and we would like to obtain a fine-grained version. 
Let ${\mathcal S}$ be 
the collection of indices of auxiliary data sets. 
The notations used in this paper are listed in Table~\ref{notation}. 

{\bf Partition:}
Let ${\mathcal X}$ be a total region of a city, 
and ${\vect x} \in {\mathcal X}$ be a location point
represented by its coordinates (e.g., latitude and longitude).
Partition ${\mathcal P}$ of ${\mathcal X}$ is
a collection of disjoint subsets, 
called {\em regions}, of ${\mathcal X}$,
whose union is equal to ${\mathcal X}$.
Let $|{\mathcal P}|$ denote the number of regions in ${\mathcal P}$. 
We can consider several partitions of ${\mathcal X}$ as follows. 
Let ${\mathcal P}^{\rm coar}$ be the coarse-grained partition, 
i.e., that of the coarse-grained target data. 
Let ${\mathcal P}^{\rm fine}$ be the fine-grained partition, 
of the desired fine-grained target data. 
For $s\in{\mathcal S}$, let ${\mathcal P}_s$ be the partition of the $s$th auxiliary data set. 

{\bf Spatial data:}
Let ${\vect a} = \T{(a_1,\ldots,a_{|{\mathcal P}^{\rm coar}|})}$ 
be a $|{\mathcal P}^{\rm coar}|$-dimensional vector consisting of the coarse-grained target values,
where $a_i \in {\mathbb R}$ is the value associated with region $i \in {\mathcal P}^{\rm coar}$.
For $s\in{\mathcal S}$, let ${\vect y}_s = \T{(y_{s,1},\ldots,y_{s,|{\mathcal P_s}|})}$ be a $|{\mathcal P_s}|$-dimensional vector 
consisting of the $s$th auxiliary data values, where $y_{s,p} \in {\mathbb R}$ is the value associated 
with region $p \in {\mathcal P_s}$ of the $s$th auxiliary data set. 

{\bf Problem:}
Suppose that we have coarse-grained target data ${\vect a}$ whose partition is ${\mathcal P}^{\rm coar}$, 
auxiliary data sets with the respective partitions $\{ ({\mathcal P}_s,{\vect y}_s) \mid s \in {\mathcal S} \}$, 
and the desired fine-grained partition ${\mathcal P}^{\rm fine}$, 
we wish to estimate a $|{\mathcal P}^{\rm fine}|$-dimensional vector 
${\vect z} = \T{(z_1,\ldots,z_{|{\mathcal P}^{\rm fine}|})}$ 
consisting of the fine-grained target values, where $z_j \in {\mathbb R}$ is the value associated with region $j \in {\mathcal P}^{\rm fine}$. 
Here, the values $a_i$, $y_{s,p}$ and $z_j$ are assumed to be intensive quantities such as ratios; that is, 
they are independent of the area scale of the respective regions. 
When the values are extensive quantities such as population, 
they can be transformed into intensive quantities 
by dividing them with the areas of regions.

\section{Proposed Model}
\label{sec:proposed}
\begin{figure}
 \begin{center}
  \includegraphics[width=60mm]{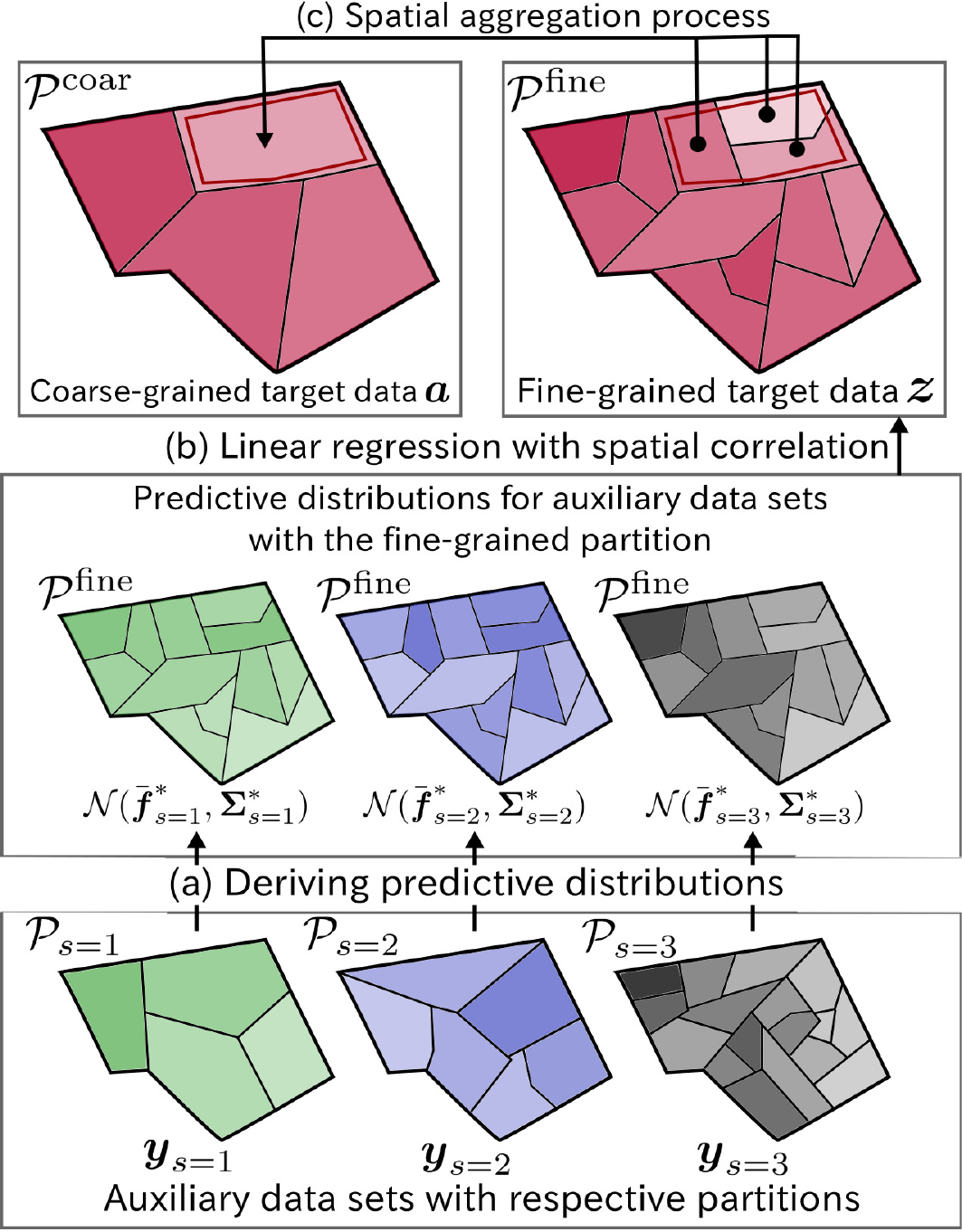}
 \end{center}
 \vspace{-15pt}
 \caption{Generative process of coarse-grained target data given three auxiliary data sets.}
 \label{fig:model}
 \vspace{-12pt}
\end{figure}
We propose a probabilistic model that allows auxiliary spatial data sets 
with various granularities to be used in refining coarse-grained spatial data. 
Our model is based on Gaussian process (GP)~\cite{carl:gaussian}, 
which is a flexible non-parametric model for non-linear functions in a continuous domain. 
We model the generative process for coarse-grained target data ${\vect a}$, 
given the auxiliary data sets with known partitions
$\{ ({\mathcal P}_s,{\vect y}_s) \mid s \in {\mathcal S} \}$, 
coarse-grained partition ${\mathcal P}^{\rm coar}$, 
and fine-grained partition ${\mathcal P}^{\rm fine}$. 
In other words,
we model the conditional probability
$p({\vect a} \mid \{{\vect y}_s\}_{s \in {\mathcal S}})$ 
instead of the joint probability of ${\vect a}$ and $\{{\vect y}_s \}_{s \in {\mathcal S}}$.
It enables us to adopt two-step inference approach
described in Section~\ref{sec:inference}, 
which is advantageous in the computational cost for learning model parameters. 

The generative process (given three auxiliary data sets) is illustrated schematically 
in Figure~\ref{fig:model}, where darker hues represent regions with higher values.
This process contains the following three steps:
(a) Deriving the predictive distribution over continuous space 
for each auxiliary data set ${\vect y}_s$ via GP regression, 
which corresponds to spatial interpolation;
(b) generating the fine-grained target data ${\vect z}$ via a GP 
whose mean function is modeled as the linear regression 
of the continuous predictive distributions of the auxiliary data sets; 
(c) generating the coarse-grained target data ${\vect a}$ 
by spatially aggregating the constituent values in a fine-grained partition. 

In our problem, 
each value is associated with a region in a partition rather than a single location point in ${\mathcal X}$;
this prevents us from directly applying GP.
We thus associate each region in a partition with its centroid,
and regard each value as being associated with the centroid of that region.
This assumption, while significantly simplifying computations involved,
might worsen the fit of the GP to the data set,
which however is appropriately taken into account in the following steps
as increased uncertainty of the
GPs for both the respective auxiliary data sets (described in~(\ref{eq:marginal_y}))
and the target data (described in~(\ref{eq:marginal_a})).
For $s\in{\mathcal S}$,
let ${\bf X}_s = ({\vect x}_{s,1},\ldots,{\vect x}_{s,|{\mathcal P_s}|})$ be 
the set of the centroids in partition ${\mathcal P}_s$, 
where ${\vect x}_{s,p}$ is the centroid of region $p \in {\mathcal P_s}$.
Similarly, for fine-grained partition ${\mathcal P}^{\rm fine}$, 
let ${\bf X}^{\rm fine} = ({\vect x}_1,\ldots,{\vect x}_{|{\mathcal P}^{\rm fine}|})$ be the set of centroids in ${\mathcal P}^{\rm fine}$.
Thus, our problem is now reformulated as estimating ${\vect z} = \T{(z_1,\ldots,z_{|{\mathcal P}^{\rm fine}|})}$, 
where $z_j \in {\mathbb R}$ is a target value at the centroid of region $j \in {\mathcal P}^{\rm fine}$, 
as indicated by the auxiliary spatial data sets $\{({\bf X}_s,{\vect y}_s)\mid s\in{\mathcal S}\}$. 

{\bf (a) Deriving predictive distributions of auxiliary spatial data sets:}
In order to handle auxiliary spatial data sets with various granularities,
we use GP regression to derive a posterior Gaussian process for a latent continuous random function on $\mathcal{X}$; 
this conceptually corresponds to spatial interpolation of each auxiliary spatial data set. 
We then evaluate the predictive distribution on the basis of the posterior Gaussian process. 
Let $f_s({\vect x})$ be a noise-free latent function for the $s$th auxiliary data set at location ${\vect x}$.
We assume that $f_s({\vect x})$ follows a Gaussian process, 
$f_s({\vect x}) \sim \mathcal{GP} (0, k_s({\vect x}, {\vect x}^\prime))$, 
with mean zero and 
a covariance function $k_s({\vect x}, {\vect x}^\prime)$.
Though our model does not depend on any particular choice of the covariance function,
for simplicity we consider the well-known covariance function, 
i.e., squared-exponential kernel, 
which is widely used for measuring the similarity between function values 
in spatial coordinates~\cite{carl:gaussian}. 
The squared-exponential kernel is defined as
\vspace{-0.15\baselineskip}
\begin{equation}
 k_s({\vect x}, {\vect x}^\prime) 
  = \alpha^2_s \exp \left( 
		     -\frac{1}{2\gamma^2_s} \|{\vect x} - {\vect x}^\prime\|^2 
		    \right), 
  \label{eq:kernel}
\end{equation}
where $\gamma_s$ is the scale parameter, 
$\alpha^2_s$ is a signal variance that controls the magnitude of the covariance, 
and $\|\cdot\|$ is the Euclidean norm. 
We assume that the $s$th auxiliary data ${\vect y}_s$ is generated with an additive Gaussian noise with noise variance $\sigma_s^2$. 
Defining $f_s^\ast({\vect x}_p)$ as
the prediction of the $s$th
auxiliary data set for the centroid ${\vect x}_p$ of the 
fine-grained partition,
the predictive distribution of 
${\vect f}^\ast_s = 
(f_s^\ast({\vect x}_1),\ldots,f_s^\ast({\vect x}_{|{\mathcal P}^{\rm fine}|}))^\top$ 
is as follows:
\vspace{-0.15\baselineskip}
\begin{equation}
 p({\vect f}_s^\ast) = 
  {\mathcal N}({\vect f}_s^\ast \mid \bar{{\vect f}_s^\ast}, {\vect \Sigma}_s^\ast),
  \label{eq:prediction_s}
\end{equation}
where $\bar{{\vect f}_s^\ast} = 
\T{{\bf K}_{s\ast}} ({\bf K}_s + \sigma_s^2 {\bf I})^{-1} {\vect y}_s$ is 
the predictive means, 
and ${\vect \Sigma}_s^\ast = {\bf K}_{s\ast\ast} 
 - \T{{\bf K}_{s\ast}} ({\bf K}_s + \sigma_s^2 {\bf I})^{-1} {\bf K}_{s\ast}$ is 
the covariance matrix, 
whose diagonal elements represent the uncertainties in the prediction at the test points ${\bf X}^{\rm fine}$. 
Incorporation of 
the predictive distributions~(\ref{eq:prediction_s}) is expected 
to allow the usefulness of auxiliary data to be effectively learnt
as it allows consideration of the uncertainty in the prediction. 
Details are given in~(\ref{eq:lambda}) in Section~\ref{sec:inference}. 
Here, 
${\bf K}_s$ is 
a $|{\mathcal P}_s| \times |{\mathcal P}_s|$ covariance matrix 
whose entries are covariances 
between training points ${\bf X}_s$. 
${\bf K}_{s\ast}$ is 
a $|{\mathcal P}_s| \times |{\mathcal P}^{\rm fine}|$ covariance matrix 
whose entries are covariances 
between training points ${\bf X}_s$ and test points ${\bf X}^{\rm fine}$. 
${\bf K}_{s\ast\ast}$ is 
a $|{\mathcal P}^{\rm fine}| \times |{\mathcal P}^{\rm fine}|$ covariance matrix 
whose entries are covariances 
between test points ${\bf X}^{\rm fine}$. 

{\bf (b) Generative process of fine-grained target data:}
We model a generative process 
for the fine-grained target data ${\vect z}$. 
Let $z({\vect x})$ be 
a noise-free latent function 
for the fine-grained target data
at location ${\vect x}$. 
We assume that $z({\vect x})$ follows a Gaussian process, 
${\vect z}({\vect x}) \sim \mathcal{GP} (m({\vect x}), k({\vect x}, {\vect x}^\prime))$,
with mean function $m({\vect x}) = \sum_{s \in {\mathcal S}} w_s f_s({\vect x}) + w_0$, 
where $w_s \in {\mathbb R}$ and $w_0 \in {\mathbb R}$ are 
the regression coefficient of the $s$th auxiliary data set and the bias parameter, respectively. 
The covariance function $k({\vect x}, {\vect x}^\prime)$ 
is a squared-exponential kernel 
with the scale parameter $\gamma$ and signal variance $\alpha^2$. 
Given the predictive values for the auxiliary data sets from~(\ref{eq:prediction_s}), 
the conditional distribution of ${\vect z}$ at the centroids ${\bf X}^{\rm fine}$ is given by
\vspace{-0.15\baselineskip}
\begin{equation}
 p({\vect z} \mid {\bf F}^*) = 
  {\mathcal N}({\vect z}\mid{\bf F}^\ast{\vect w}, {\bf K}), 
  \label{eq:generate_z}
\end{equation}
where ${\vect w} = \T{(w_0,\ldots,w_{|{\mathcal S}|})}$ and ${\bf K}$ is 
a $|{\mathcal P}^{\rm fine}| \times |{\mathcal P}^{\rm fine}|$ covariance matrix
defined by $k({\vect x}, {\vect x}^\prime)$.
Here, we let $|{\mathcal S}|$ be the number of auxiliary data sets. 
We define the augmented matrix as 
the $|{\mathcal P}^{\rm fine}| \times (|{\mathcal S}| + 1)$ matrix 
${\bf F}^\ast = ({\vect f}_1^\ast,\ldots,{\vect f}_{|{\mathcal S}|}^\ast,{\bf 1})$, 
in which ${\vect 1}$ is a column vector of 1's. 
This GP-based modeling enables us to consider the spatial correlation 
in the target data and the auxiliary data sets simultaneously.

{\bf (c) Generative process of coarse-grained target data:}
We design a spatial aggregation process to transform the fine-grained target data ${\vect z}$ into the coarse-grained target data ${\vect a}$,
in order to encourage consistency between ${\vect z}$ , which is to be estimated, and the available coarse-grained target data ${\vect a}$. 
In the spatial aggregation process, 
a value associated with one region in the coarse-grained partition is obtained by aggregating the values in the fine-grained regions contained in the coarse-grained region
(see the upper part of Figure~\ref{fig:model}). 
Then, ${\vect a}$ is generated from the following conditional distribution given ${\vect z}$,
\vspace{-0.15\baselineskip}
\begin{equation}
 p({\vect a} \mid {\vect z}) = 
  \mathcal{N}({\vect a} \mid {\bf H}{\vect z}, \sigma^2{\bf I}),
  \label{eq:generate_a}
\end{equation}
where $\sigma^2$ is 
the noise variance for the coarse-grained target data,
and ${\bf H}$ is 
a $|{\mathcal P}^{\rm coar}| \times |{\mathcal P}^{\rm fine}|$ aggregation matrix,
whose entries are nonnegative weighting coefficients; 
the row sum of ${\bf H}$ should equal 1. 
We set the coefficients in accordance with the property of the target data. 
For example,
in cases where target data are incidences of disease,
then the $(i,j)$-entry $H(i,j)$ of ${\bf H}$
would be proportional to the population in the intersection of the coarse-grained region $i$ and the fine-grained region $j$.
In the following, for simplicity, 
we consider a simple aggregation matrix, 
in which entry $H(i,j)$ is 
$1 / |{\mathcal P}^{\rm fine}_i|$ 
if the fine-grained region $j$ is 
contained in the coarse-grained region $i$, and zero otherwise.
Here, ${\mathcal P}^{\rm fine}_i$ is 
a subset of ${\mathcal P}^{\rm fine}$, all the elements of which are contained in the coarse-grained region $i\in{\mathcal P}^{\rm coar}$.

\section{Inference}
\label{sec:inference}
Given the coarse-grained target data ${\vect a}$, the auxiliary spatial data sets with centroids $\{ ({\bf X}_s, {\vect y}_s) \mid s \in {\mathcal S} \}$,
the centroids of fine-grained partition ${\bf X}^{\rm fine}$ and the aggregation matrix ${\bf H}$, we aim to predict the fine-grained target data ${\vect z}$ 
via a Bayesian inference procedure. 
In order to calculate 
the predictive distribution of ${\vect z}$, 
we need to estimate the model parameters. 
The problem of estimating the model parameters can be divided into two steps:
1) Estimate hyperparameters $\alpha_s, \gamma_s, \sigma_s$ 
for each auxiliary data set
and 2) estimate regression coefficient ${\vect w}$ 
and hyperparameters $\alpha, \gamma, \sigma$ 
for the target data. 
Although one could also opt for estimating all the model parameters 
simultaneously (i.e., one-step inference),
it will increase the computational cost of inference drastically;
we adopt the efficient two-step inference as described in the following paragraphs.
We finally construct 
the predictive distribution of ${\vect z}$ 
by using the estimated parameters.
Details of the inference procedure are shown in Algorithm~\ref{alg1}.

\setlength{\textfloatsep}{7pt}{
\begin{algorithm}[t]
 \caption{Bayesian inference procedure of the fine-grained target data ${\vect z}$}
 \label{alg1}
 \SetKwInOut{Input}{Input}
 \SetKwInOut{Output}{Output}
 \Input{${\vect a}$, $\{ ({\bf X}_s, {\vect y}_s) \mid s \in {\mathcal S} \}$, 
 ${\bf X}^{\rm fine}$, ${\bf H}$}
 \Output{Predictive distribution of ${\vect z}$}
 \begin{algorithmic}[1]
  \STATE Initialize model parameters, 
  $\{\alpha_s \mid s \in {\mathcal S}\}$, $\{\gamma_s \mid s \in {\mathcal S}\}$, 
  $\{\sigma_s \mid s \in {\mathcal S}\}$, ${\vect w}$, $\alpha$, $\gamma$, $\sigma$
  \STATE /* first inference step */
  \STATE {\bf for} $s \in {\mathcal S}$ {\bf do}
  \STATE \quad Estimate $\alpha_s, \gamma_s, \sigma_s$ 
  by maximizing the logarithm of~(\ref{eq:marginal_y})
  \STATE {\bf end for}
  \STATE /* second inference step */
  \STATE Estimate ${\vect w}, \alpha, \gamma, \sigma$ 
  by maximizing the logarithm of~(\ref{eq:marginal_a})
  \STATE Construct predictive distribution of ${\vect z}$ by (\ref{eq:predictive_z})
  using the estimated model parameters
 \end{algorithmic}
\end{algorithm}}

{\bf The first inference step:}
Given the $s$th auxiliary spatial data set with centroids $({\bf X}_s, {\vect y}_s)$,
the marginal likelihood of ${\vect y}_s$ is given by
\vspace{-0.15\baselineskip}
\begin{equation}
p({\vect y}_s | \alpha_s, \gamma_s, \sigma_s) = 
{\mathcal N}({\vect y}_s | 0, {\bf K}_s + \sigma_s^2 {\bf I}).
\label{eq:marginal_y}
\end{equation}
The hyperparameters $\alpha_s, \gamma_s, \sigma_s$ are estimated 
by maximizing the logarithm of~(\ref{eq:marginal_y}).
We solve the optimization problem through the use of the BFGS method~\cite{liu:on}. 
By solving the optimization problem for each auxiliary data set independently, 
we obtain the set of the estimated hyperparameters for all auxiliary data sets.
The predictive distribution of ${\vect f}_s^\ast$ corresponding to~(\ref{eq:prediction_s}) is obtained using the estimated hyperparameters.

{\bf The second inference step:}
Given the coarse-grained target data ${\vect a}$ 
and the centroids of fine-grained partition ${\bf X}^{\rm fine}$, 
the marginal likelihood of ${\vect a}$ is given by
\vspace{-0.15\baselineskip}
{\small
\begin{flalign}
 p({\vect a} \mid {\vect w}, \alpha, \gamma, \sigma) 
 &= \int \int p({\vect a} \mid {\vect z}) p({\vect z} \mid {\bf F}^\ast) 
 \prod_{s \in {\mathcal S}}p({\vect f}_s^\ast) d{\bf F}^\ast d{\vect z} \nonumber \\
 &= \int \int {\mathcal N}\left({\vect a}\mid{\bf H}{\vect z}, \sigma^2 {\bf I}\right) 
 {\mathcal N}\left({\vect z}\mid{\bf F}^\ast {\vect w}, {\bf K}\right) \nonumber \\
 &\quad\quad\quad \times \prod_{s \in {\mathcal S}} {\mathcal N}\left({\vect f}_s^\ast\mid\bar{{\vect f}_s^\ast}, {\vect \Sigma}_s^\ast\right)
 d{\bf F}^\ast d{\vect z} \nonumber \\
 & = {\mathcal N} \left({\vect a}\mid{\bf H} \bar{{\bf F}}^\ast {\vect w}, {\vect \Lambda} \right),
 \label{eq:marginal_a}
\end{flalign}
}where 
$\bar{{\bf F}}^\ast = 
(\bar{{\vect f}_1^\ast},\ldots,\bar{{\vect f}}^\ast_{|{\mathcal S}|}, {\bf 1})$
is a $|{\mathcal P}^{\rm fine}| \times (|{\mathcal S}| + 1)$ matrix, 
and we analytically integrate out the latent variables ${\bf F}^\ast$ and ${\vect z}$ 
with the help of the conjugacy 
of the distributions~(\ref{eq:prediction_s}), (\ref{eq:generate_z}), 
and (\ref{eq:generate_a}). 
${\vect \Lambda}$ is a $|{\mathcal P}^{\rm coar}| \times |{\mathcal P}^{\rm coar}|$ 
covariance matrix represented by 
${\vect \Lambda} = \sigma^2 {\bf I} + {\bf H} {\bf \Omega} \T{{\bf H}}$,
where ${\bf \Omega} = {\bf K} + \sum_{s \in {\mathcal S}}w_s^2 {\vect \Sigma}_s^\ast$.
The $(i,i^\prime)$-entry $\Lambda(i,i^\prime)$ of ${\vect \Lambda}$
is shown in~(\ref{eq:lambda}).
\begin{table*}[b]
 \small
 \centering
 \hrulefill
 \vspace{-0.5\baselineskip}
 \begin{flalign}
  \hspace{-5pt}
   \Lambda(i,i^\prime) =& \sigma^2 \delta_{i,i^\prime} 
  + \frac{1}{|{\mathcal P}^{\rm fine}_i||{\mathcal P}^{\rm fine}_{i^\prime}|}
  \nonumber \\
  &\times
   \sum_{j \in {\mathcal P}^{\rm fine}_i} 
   \sum_{j^\prime \in {\mathcal P}^{\rm fine}_{i^\prime}}
   \Biggl[
   \delta_{j,j^\prime}
   \underbrace{
   \left(
    \alpha^2 + \sum_{s \in {\mathcal S}} w_s^2 \Sigma_s^\ast(j,j^\prime)
	 \right)
   }_{\text{residual variance term}}
   +
   (1-\delta_{j,j^\prime})
   \underbrace{
   \left(
    \alpha^2 \exp
    \left(-\frac{1}{2\gamma^2} \|{\vect x}_j - {\vect x}_{j^\prime}\|^2\right) 
    + \sum_{s \in {\mathcal S}} w_s^2 \Sigma_s^\ast(j,j^\prime)
	\right)
   }_{\text{spatial correlation term}}
   \Biggr] 
   \label{eq:lambda}
 \end{flalign}
\end{table*}
Here, 
$\delta_{\bullet,\bullet}$ in ({\ref{eq:lambda}}) represents Kronecker delta; 
$\delta_{A,B}=1$ if $A=B$, and $\delta_{A,B}=0$ otherwise.
The residual variance term in~(\ref{eq:lambda}) represents 
the residual variance in the regression of $z({\vect x}_j)$. 
This term contains 
the uncertainty in the prediction of $f_s({\vect x}_j)$, 
i.e., $\Sigma_s^\ast(j,j)$, 
which is weighted by $w_s^2$.
The spatial correlation term in ~(\ref{eq:lambda}) represents 
the strength of spatial correlation 
between $z({\vect x}_j)$ and $z({\vect x}_{j^\prime})$. 
This term contains 
the covariance between $f_s({\vect x}_j)$ and $f_s({\vect x}_{j^\prime})$, 
i.e., $\Sigma_s^\ast(j,j^\prime)$, which is weighted by $w_s^2$.
On the basis of the marginal likelihood~(\ref{eq:marginal_a}) 
with this covariance matrix ${\bf \Lambda}$,
our model can effectively learn the regression coefficient ${\vect w}$ 
while taking into consideration the prediction uncertainties and the spatial correlations from the auxiliary data sets with various granularities, simultaneously. 
The parameter ${\vect w}$ and the hyperparameters $\alpha$, $\gamma$, $\sigma$ 
are estimated by maximizing 
the logarithm of (\ref{eq:marginal_a}). 
We solve the optimization problem by using the BFGS method~\cite{liu:on}. 
The derivatives of the logarithm of (\ref{eq:marginal_a}) 
with respect to $w_s$, $\alpha$, $\gamma$, $\sigma$ are described 
in Appendix~\ref{sec:derivative}.

{\bf Predictive distribution of fine-grained target data:}
Using the estimated model parameters, the predictive distribution of the fine-grained target data ${\vect z}$ is given by
\vspace{-0.15\baselineskip}
\begin{equation}
 p({\vect z}^\ast) = 
  {\mathcal N}\left(
	       {\vect z}^\ast\mid\bar{{\vect z}}^\ast, {\rm cov}({\vect z}^\ast)
	      \right),
  \label{eq:predictive_z}
\end{equation}
where 
$\bar{{\vect z}}^\ast = 
\bar{{\bf F}}^\ast {\vect w} + {\vect \Omega}\T{{\bf H}} {\vect \Lambda}^{-1} 
({\vect a}-\bar{{\bf F}}^\ast {\vect w})$
is the predictive means, 
and where 
${\rm cov}({\vect z}^\ast) = 
{\vect \Omega} - {\vect \Omega} \T{{\bf H}} {\vect \Lambda}^{-1} {\bf H} {\vect \Omega}$
is the covariance matrix. 
We can obtain the refinement results, i.e., the estimated fine-grained target data, by using the predictive means $\bar{{\vect z}}^\ast$. 
By analyzing the covariance matrix ${\rm cov}({\vect z}^\ast)$, we can also evaluate the confidence of the refinement results.

\section{Experiments}
\label{sec:experiments}
{\bf Data description:}
We evaluated the proposed model 
using real-world spatial data sets from NYC Open Data~\footnote{https://opendata.cityofnewyork.us}.
There are 44 data sets that contain 
a variety of categories 
such as social indicators, land use, air quality and taxi traffic. 
Each data set is associated 
with one of six geographical partitions, i.e.,
school district (32), UHF42 (42), community district (59), 
police precinct (77), zip code (186) and taxi zone (249), 
where each number in parenthesis denotes the number of regions in the corresponding partition. 
In our experiments, we try to refine the poverty rate data set and the five air pollution data sets (i.e., PM2.5, ozone, formaldehyde, benzene, elemental carbon). 
The experimental setting is as follows: 
1) Given the poverty rate data set with the borough partition ($|{\mathcal P}^{\rm coar}| = 5$), 
we would like to refine the data into the community district partition ($|{\mathcal P}^{\rm fine}| = 59$), 
and 2) given each air pollution data set with the borough partition ($|{\mathcal P}^{\rm coar}| = 5$), 
we aim to refine the data into the UHF42 partition ($|{\mathcal P}^{\rm fine}| = 42$). 
Appendix~\ref{sec:discription} details the data sets and the experimental settings. 

{\bf Baselines:}
The existing methods can be applied to auxiliary data sets with various granularities 
if pre-processing is applied, i.e., spatial interpolation, 
so that the granularities of the auxiliary data sets
match with that of the fine-grained target data.
Accordingly, 
we first performed 
spatial interpolation 
of each auxiliary data set ${\vect y}_s$ by using GP regression; 
we then obtained 
the predictive values $\bar{{\vect f}^\ast_s}$ 
at the centroids ${\bf X}^{\rm fine}$ 
of the target fine-grained partition 
so that the spatial granularities of all auxiliary data sets equaled that of the fine-grained target data. 
We compared the proposed model 
with three baselines: 
GP regression (GPR)~\cite{carl:gaussian}, 
Linear regression-based method (LR-based method)~\cite{Smith:poverty}
and Two-stage statistical downscaling method (2-stage SD)~\cite{park:spatial}. 
Here, GPR is a simple spatial interpolation, 
namely, it predicts the fine-grained target data ${\vect z}$
by using only the coarse-grained target data ${\vect a}$. 
Details of these baselines are given in Appendix~\ref{sec:baselines}.

\begin{table*}[t]
\caption{MAPE $L$ and standard errors for the predictions of the fine-grained target data.}
\small
\vspace{-5pt}
\begin{center}
\begin{tabular}{@{\hspace{2.5pt}} l @{\hspace{2.5pt}} | @{\hspace{2.5pt}} c @{\hspace{2.5pt}} | @{\hspace{2.5pt}} c @{\hspace{2.5pt}} | @{\hspace{2.5pt}} c @{\hspace{2.5pt}} | @{\hspace{2.5pt}} c @{\hspace{2.5pt}} | @{\hspace{2.5pt}} c @{\hspace{2.5pt}} | @{\hspace{2.5pt}} c @{\hspace{2.5pt}}} \hline
 &PM2.5 & Ozone &Formaldehyde &Benzene &Elemental carbon &Poverty rate \\ \hline
 Proposed model
 & ${\bf 0.038 \pm 0.005}^{\star\star}$ & ${\bf 0.030 \pm 0.005}^{\star}$ & ${\bf 0.078 \pm 0.010}^{\star\star}$ & ${\bf 0.138 \pm 0.021}^{\star}$ & ${\bf 0.100 \pm 0.012}^{\star}$ & ${\bf 0.202 \pm 0.024}^{\star\star}$ \\
 2-stage SD
 & $0.052 \pm 0.007$ & $0.035 \pm 0.007$ & $0.101 \pm 0.013$ & $0.181 \pm 0.032$ & $0.123 \pm 0.016$ & $0.228 \pm 0.028$ \\
 LR-based method
 & $0.056 \pm 0.007$ & $0.040 \pm 0.007$ & $0.108 \pm 0.013$ & $0.185 \pm 0.031$ & $0.123 \pm 0.016$ & $0.234 \pm 0.028$ \\
 GPR
 & $0.072 \pm 0.010$ & $0.062 \pm 0.011$ & $0.191 \pm 0.020$ & $0.267 \pm 0.029$ & $0.195 \pm 0.019$ & $0.344 \pm 0.046$ \\
\hline
\end{tabular}
\label{MAPE}
\end{center}
\end{table*}

\begin{figure*}[!t]
\vspace{-8pt}
\begin{center}
\subfigure[True]
{\includegraphics[width=40mm]{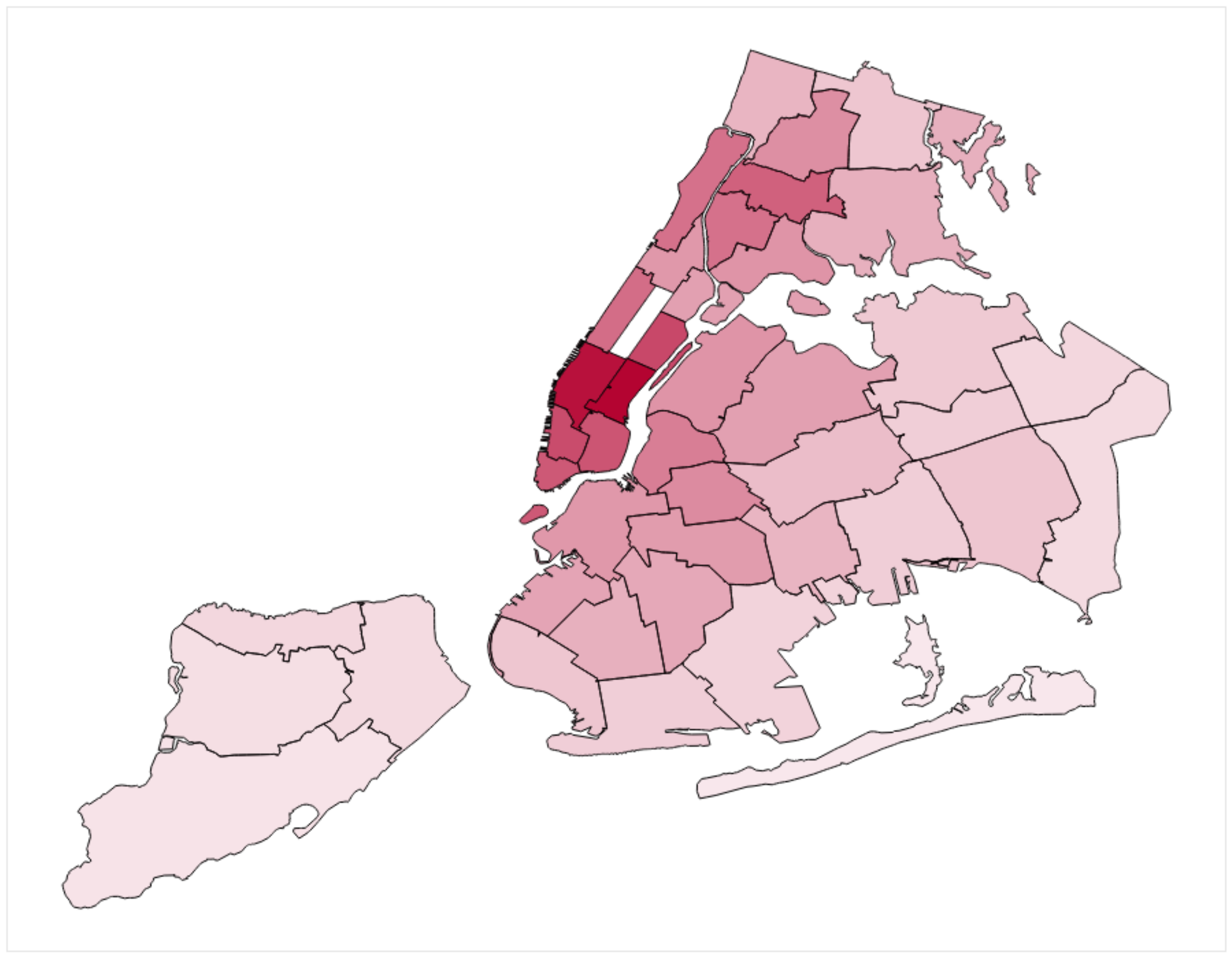}\label{fig:PM2.5_true}}
\hspace{1pt}
\subfigure[Proposed model]
{\includegraphics[width=40mm]{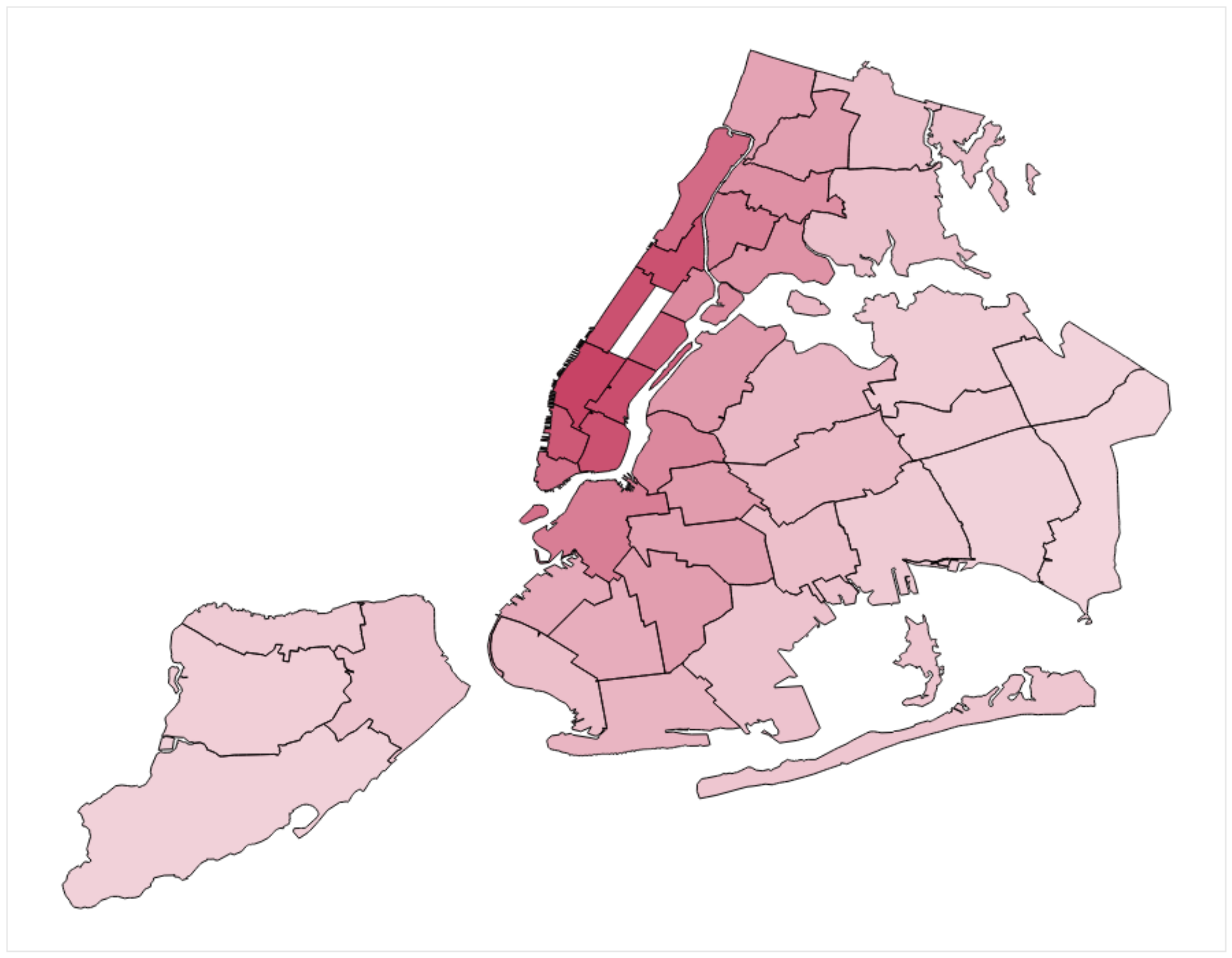}\label{fig:}}
\hspace{1pt}
\subfigure[2-stage SD]
{\includegraphics[width=40mm]{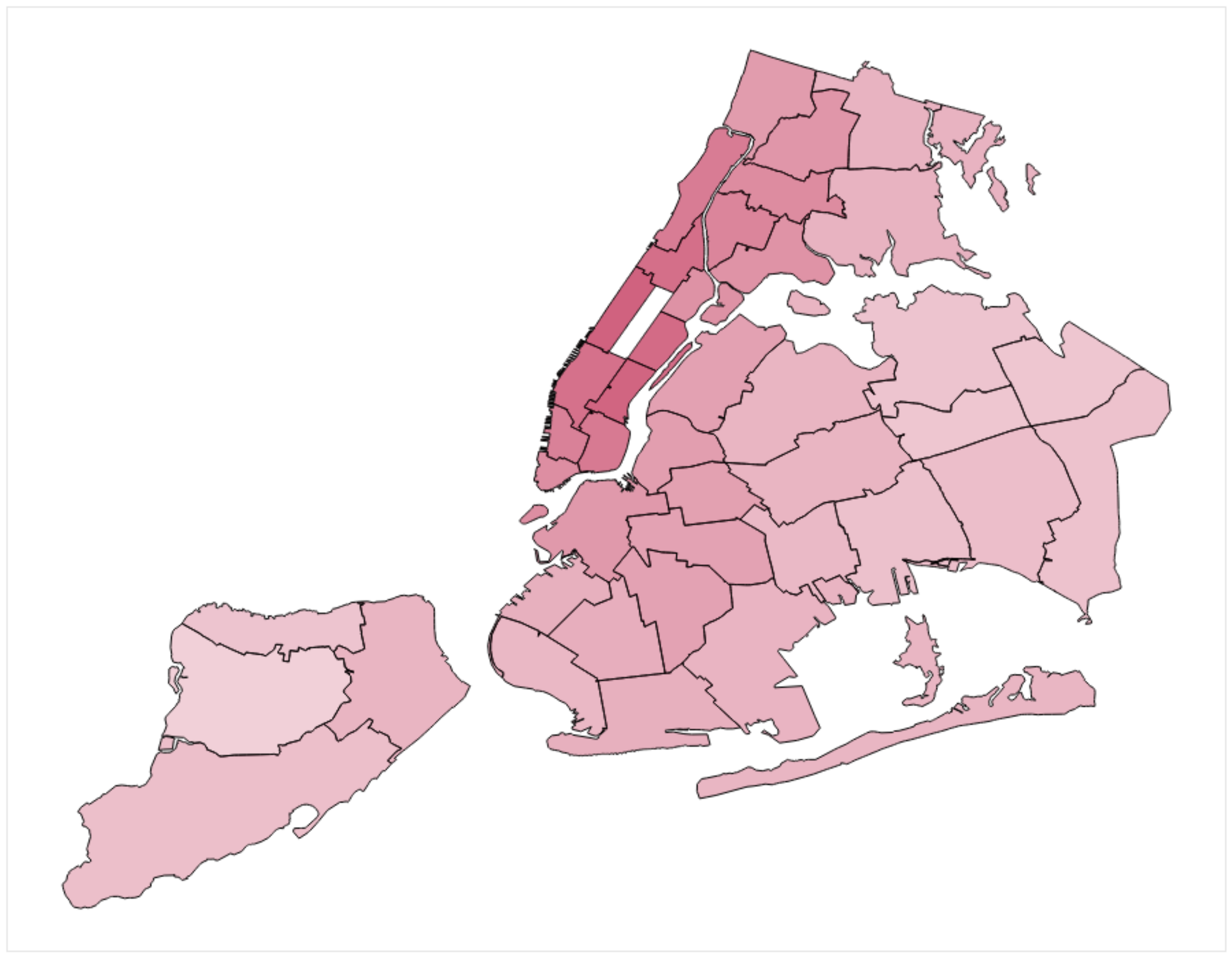}\label{fig:}}
\hspace{1pt}
\subfigure[LR-based method]
{\includegraphics[width=40mm]{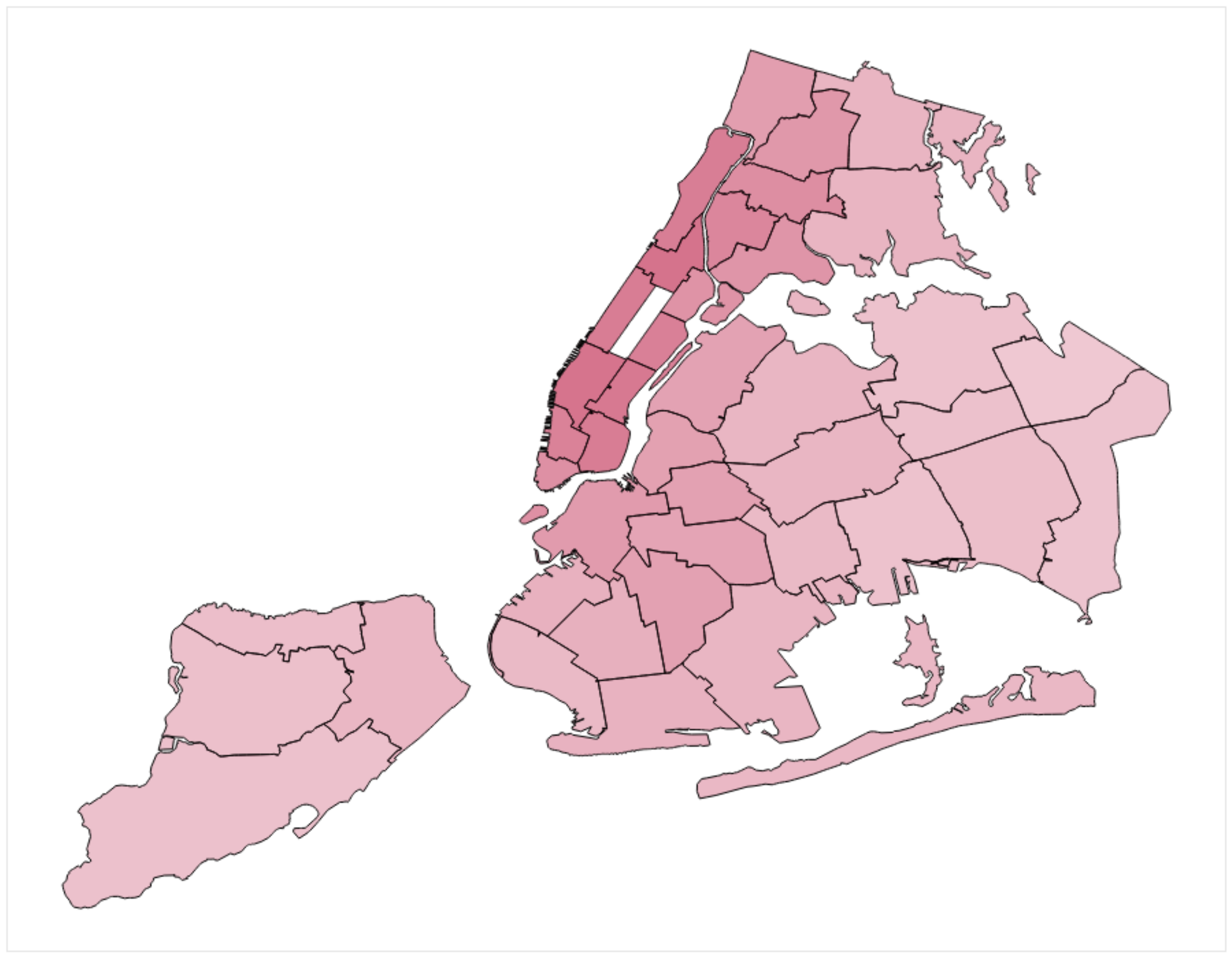}\label{fig:}}
\end{center}
\vspace{-16pt}
\caption{Comparison of the predicted fine-grained target data for PM2.5 data set.}
\label{fig:PM2.5}
\end{figure*}
\begin{figure*}[!t]
\vspace{-5pt}
\begin{center}
\subfigure[True]
{\includegraphics[width=40mm]{true_po.pdf}\label{fig:}}
\hspace{1pt}
\subfigure[Proposed model]
{\includegraphics[width=40mm]{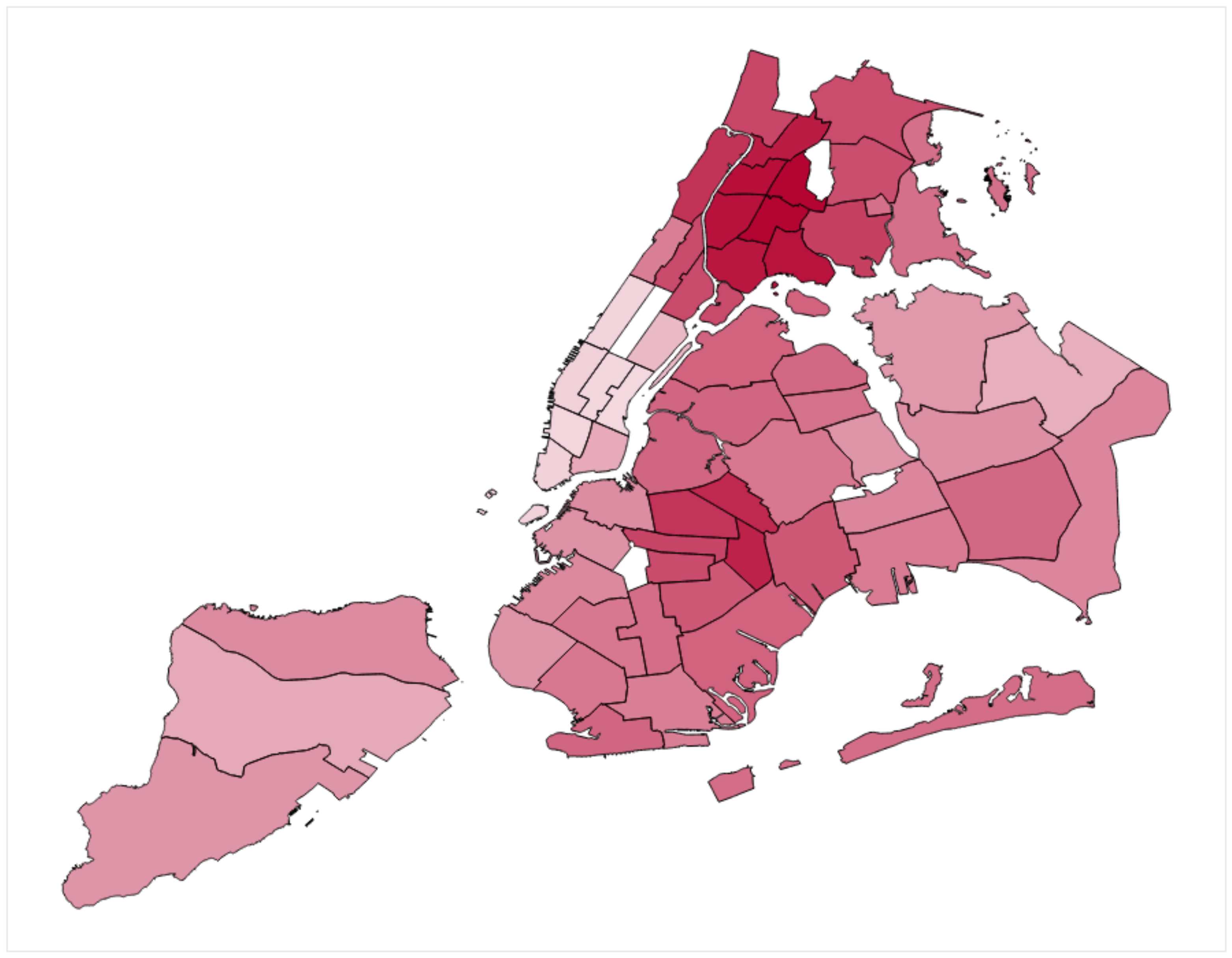}\label{fig:}}
\hspace{1pt}
\subfigure[2-stage SD]
{\includegraphics[width=40mm]{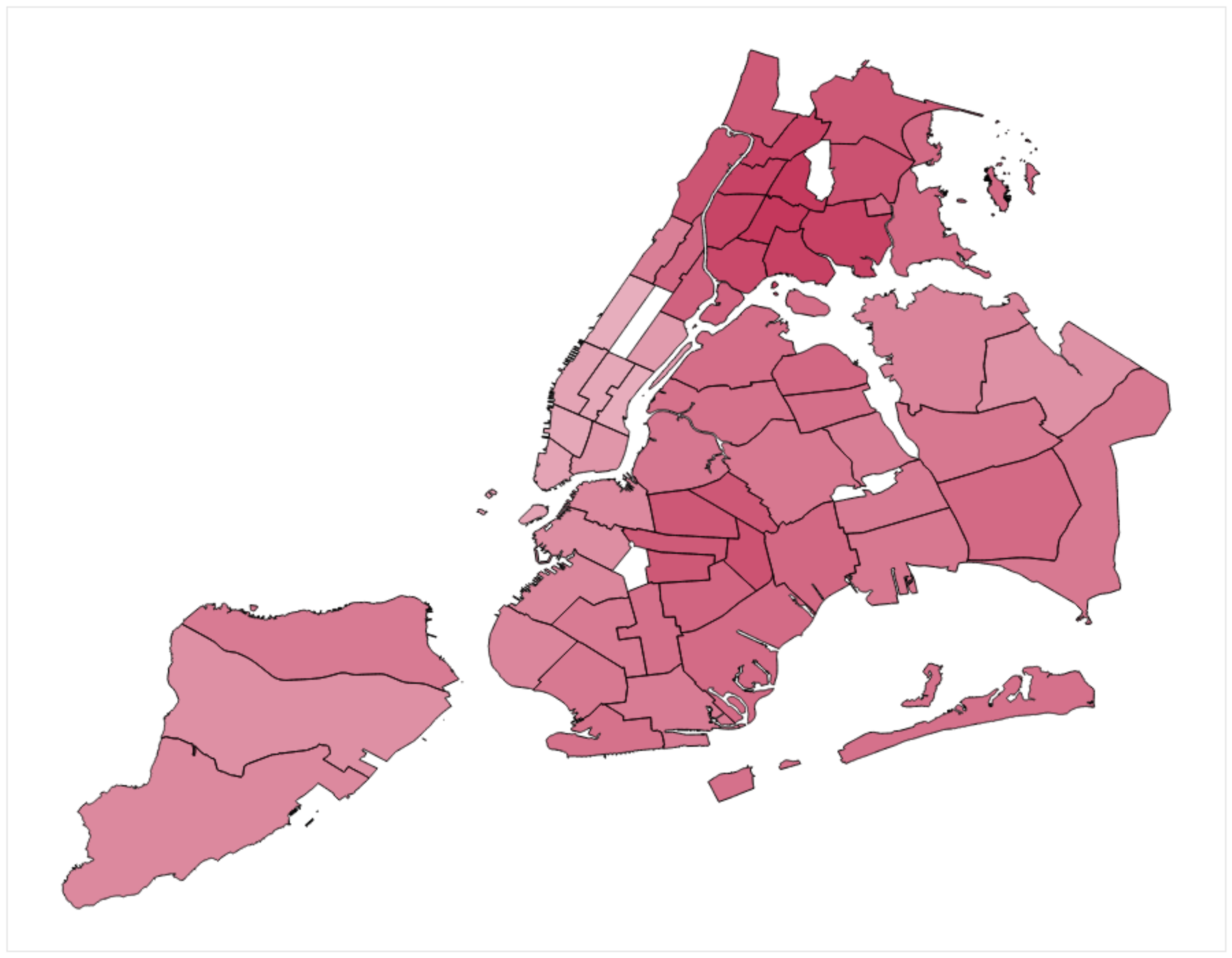}\label{fig:}}
\hspace{1pt}
\subfigure[LR-based method]
{\includegraphics[width=40mm]{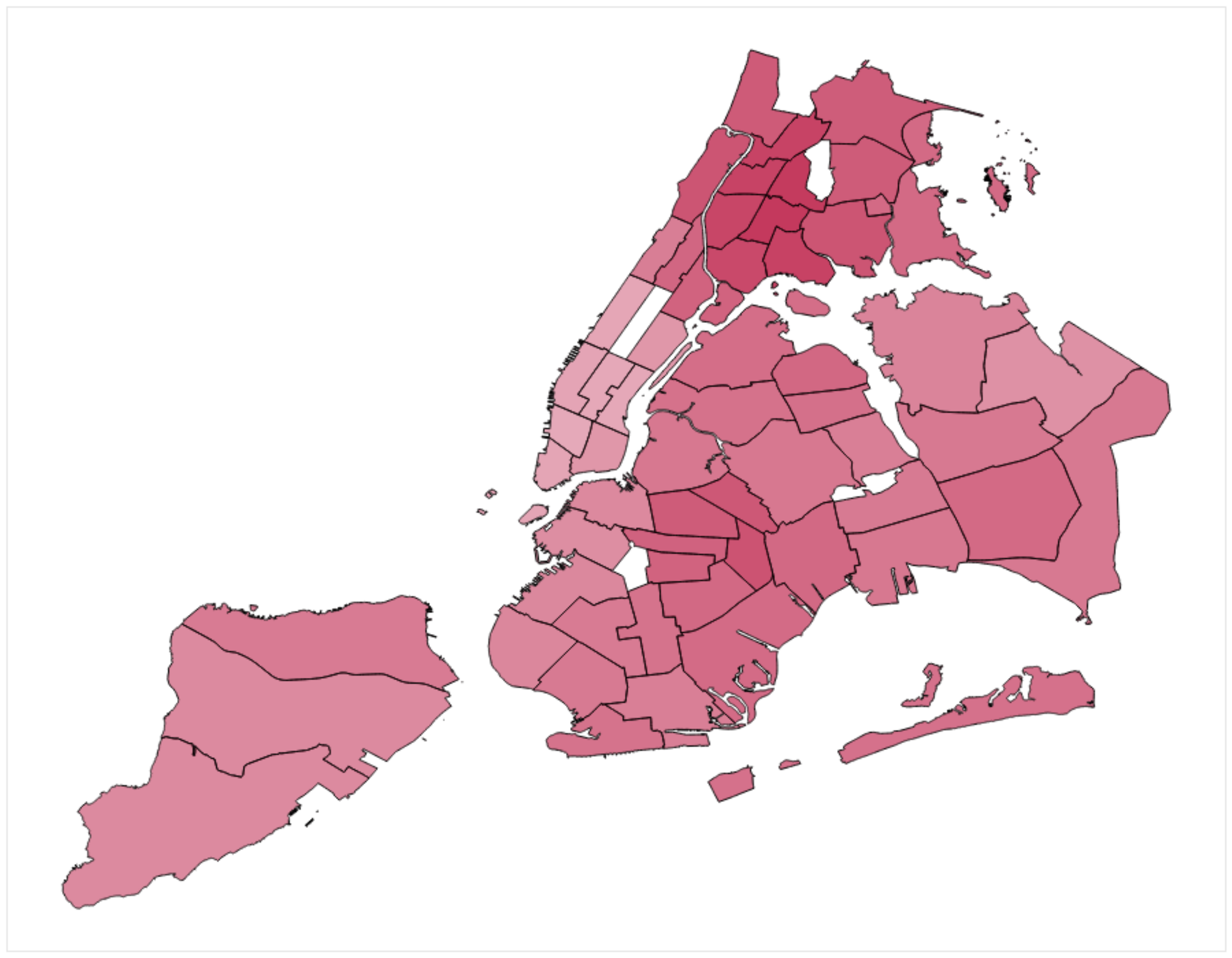}\label{fig:}}
\end{center}
\vspace{-16pt}
\caption{Comparison of the predicted fine-grained target data for poverty rate data set.}
\label{fig:poverty}
\vspace{-10pt}
\end{figure*}

{\bf Fine-grained target data prediction:} 
We evaluated our model in terms of its performance in predicting fine-grained target data ${\vect z}$.
The evaluation metric is the mean absolute percentage error (MAPE) 
in fine-grained target values: 
$L = \frac{1}{|{\mathcal P}^{\rm fine}|} \sum_{j \in {\mathcal P}^{\rm fine}}
 \left| \frac{z^{\rm true}_j - z_j^\ast}{z^{\rm true}_j} \right|$, 
where $z^{\rm true}_j$ is the true value associated with region $j$ in the target fine-grained partition;
$z_j^\ast$ is its predicted value. 
Table~\ref{MAPE} shows 
the MAPE and the standard error of absolute percentage error 
for the proposed model, 2-stage SD, LR-based method and GPR. 
For all data sets, 
our model performed better than the baselines, and the differences between our model and the baselines are statistically significant (Student's t-test).
In Table~\ref{MAPE}, 
the single star ($\star$) and the double star ($\star\star$) indicate significant difference at the levels of $P < 0.05$ and $P < 0.01$, respectively.
We found similar results using other evaluation metrics (e.g., MAE, RMSE, RMSPE). 
These results show that our model well utilized the auxiliary data sets with various granularities to accurately predict the fine-grained target data.

Figures~\ref{fig:PM2.5} and~\ref{fig:poverty} visualize 
the predicted fine-grained target data ${\vect z}$ 
for the PM2.5 data set and for the poverty rate data set, respectively. 
We illustrate the true fine-grained data 
on the left in Figures~\ref{fig:PM2.5} and~\ref{fig:poverty}, 
and the predictions made by the proposed model, 
2-stage SD and LR-based method on the right. 
Here, the predictive values of each method were normalized 
to the range $[0,1]$, 
and darker hues represent regions with higher values. 
As shown in these figures, 
our model refined 
the coarse-grained data more precisely than the other methods. 
In particular, in both data sets, 
our model achieved significant improvement 
in the north part of the map (i.e., Manhattan).
Such visualization results are useful 
for finding key regions, e.g., the poorest regions of a city. 

\begin{table}[!t]
\caption{Top-10 relevant auxiliary data as estimated by our model and 2-stage SD for PM2.5 data set.}
\scriptsize
\vspace{-3pt}
\begin{center}
\begin{tabular}{@{\hspace{1.5pt}} c @{\hspace{1.5pt}} | l@{\hspace{0.5pt}} r | l@{\hspace{0.5pt}} r} \hline
&\multicolumn{2}{c|}{Proposed model} &\multicolumn{2}{c}{2-stage SD} \\ \hline
 & Auxiliary data &$w_s$ &Auxiliary data &$w_s$ \\ \hline
1. &Fire incident (Zip code) &0.173 &1-2 fam. bldg (Comm.) &-0.088 \\
2. &Taxi dropoff (Taxi zone) &0.139 &Hospital (Comm.) &0.069 \\
3. &311 call (Zip code) &0.135 &Public school (Comm.) &0.069 \\
4. &Public telephone (Zip code) &0.114 &Lots of vacant (Comm.) &-0.067 \\
5. &Natural gas (Zip code) &0.109 &Crime (Police precinct) &0.064 \\
6. &Mean commute (Comm.) &-0.109 &Unemployment (Comm.) &0.063 \\
7. &1-2 fam. bldg (Comm.) &-0.089 &Pct. served parks (Comm.) &0.062 \\
8. &Pct. served park (Comm.) &0.075 &Library (Comm.) &0.061 \\
9. &GHG emission (Zip code) &0.068 &Fire incident (Zip code) &0.059 \\
10. &Population (Comm.) &0.062 &Park (Comm.) &0.058 \\
\hline
\end{tabular}
\label{ranking}
\end{center}
\vspace{-3pt}
\end{table}

\begin{figure}[!t]
 \centering
 \vspace{-5pt}
 \subfigure[Fire incidents]
 {\includegraphics[width=40mm]{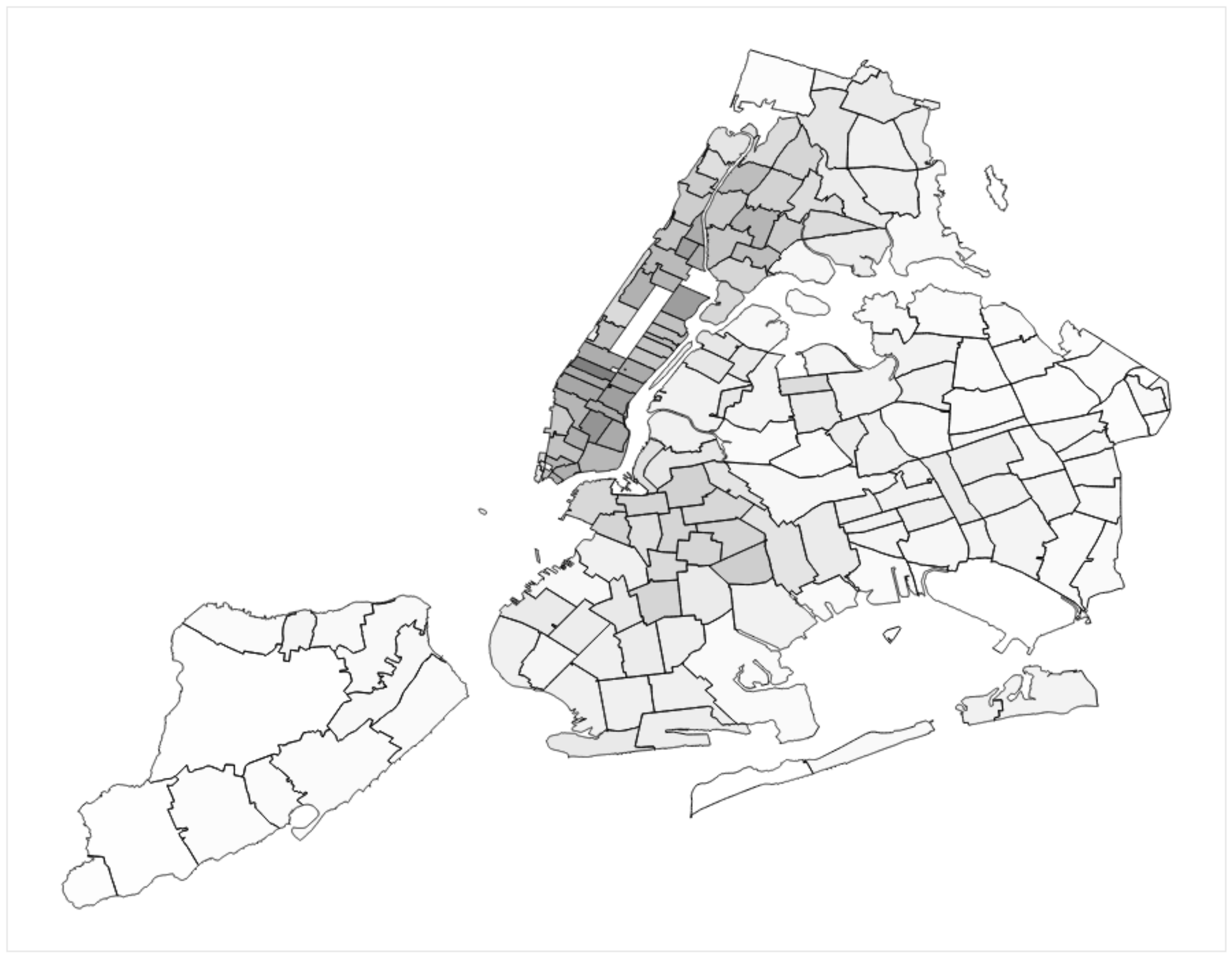}\label{fig:}}
 \hspace{1pt}
 \subfigure[Taxi dropoff]
 {\includegraphics[width=40mm]{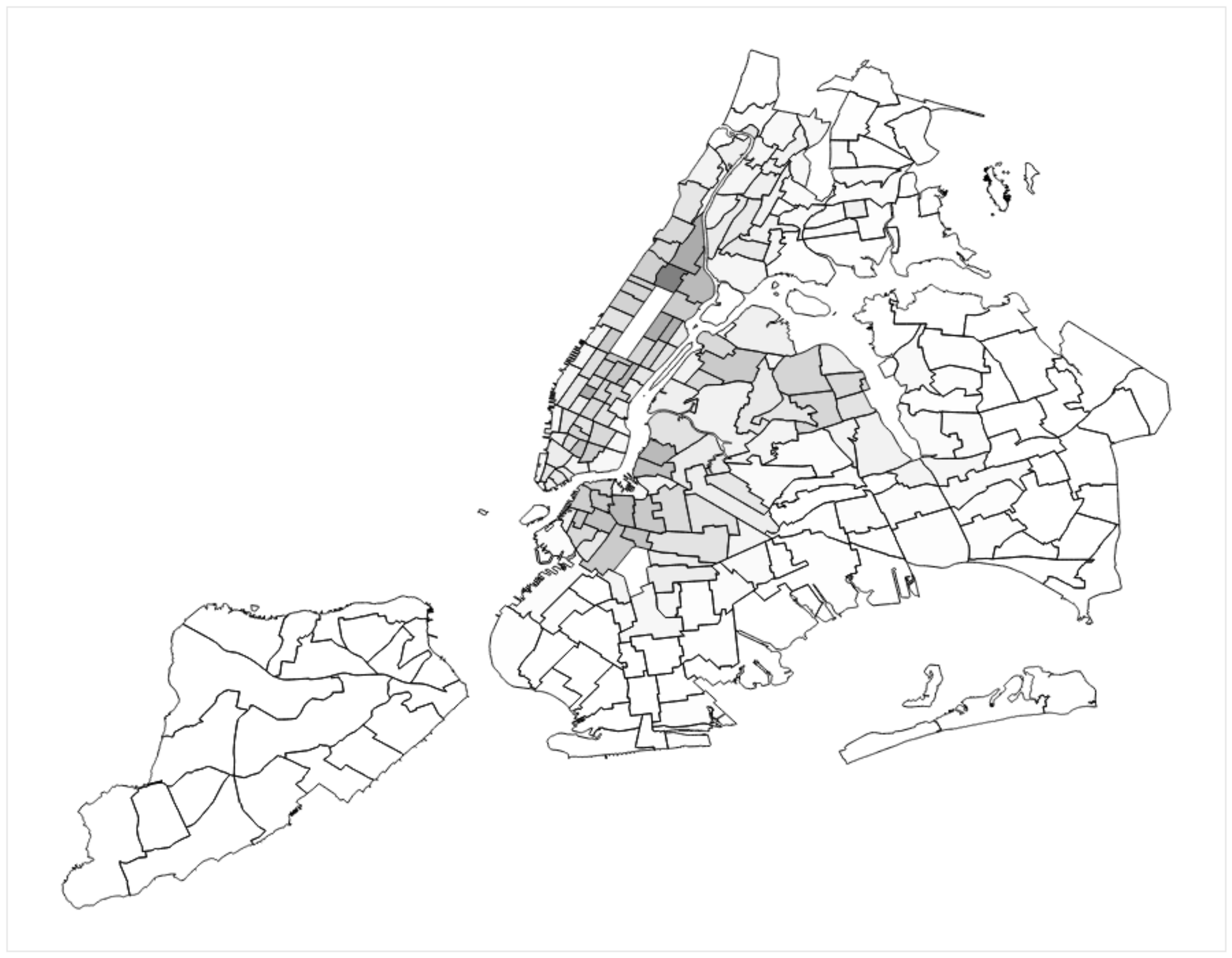}\label{fig:}}
 \vspace{-5pt}
 \caption{Top-2 auxiliary data sets ranked by the proposed model for PM2.5 data set.}
 \label{fig:top-3_P}
 \vspace{-5pt}
\end{figure}

\begin{figure}[!t]
 \centering
 \subfigure[1-2 fam. bldg]
 {\includegraphics[width=40mm]{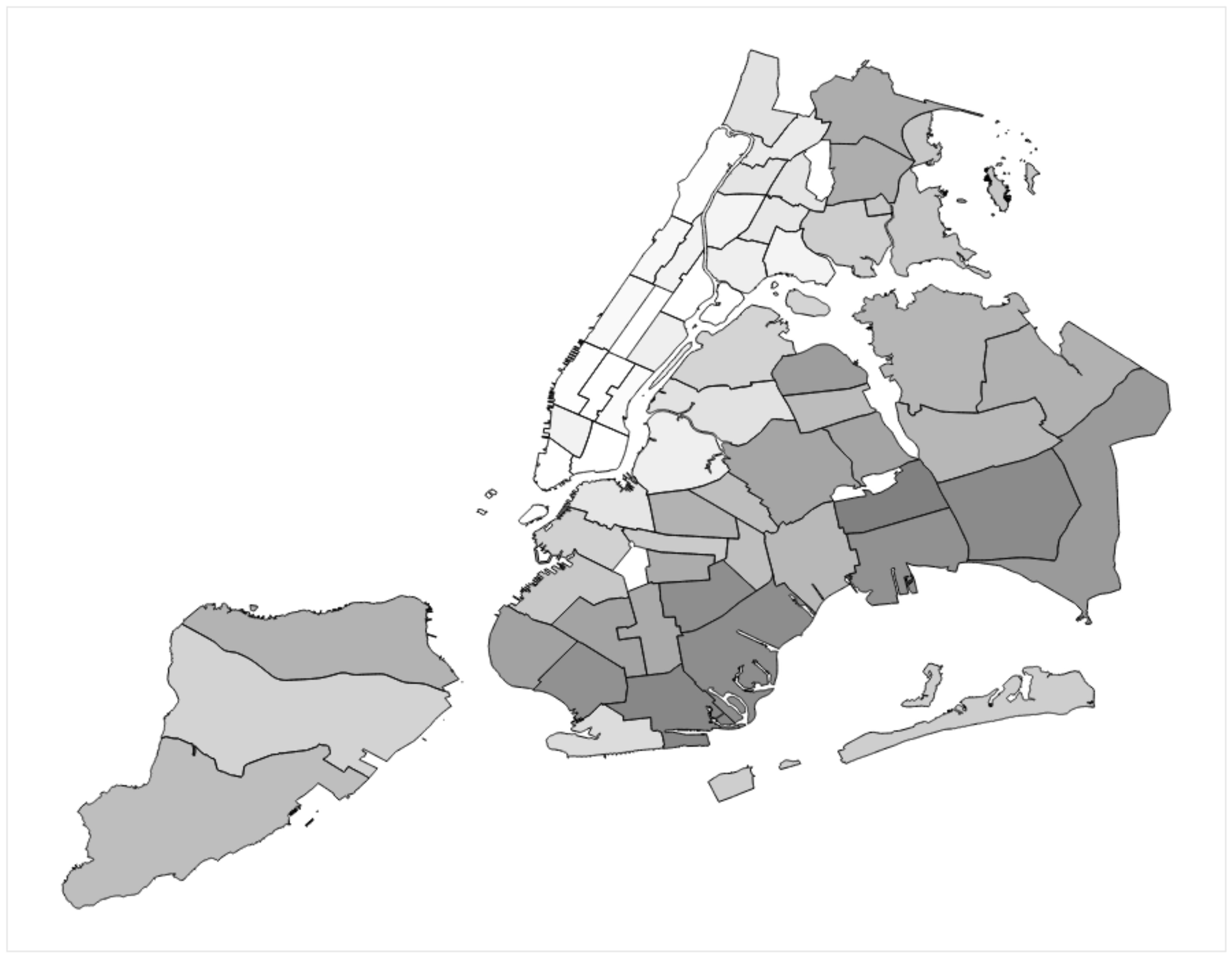}\label{fig:}}
 \hspace{1pt}
 \subfigure[Hospital]
 {\includegraphics[width=40mm]{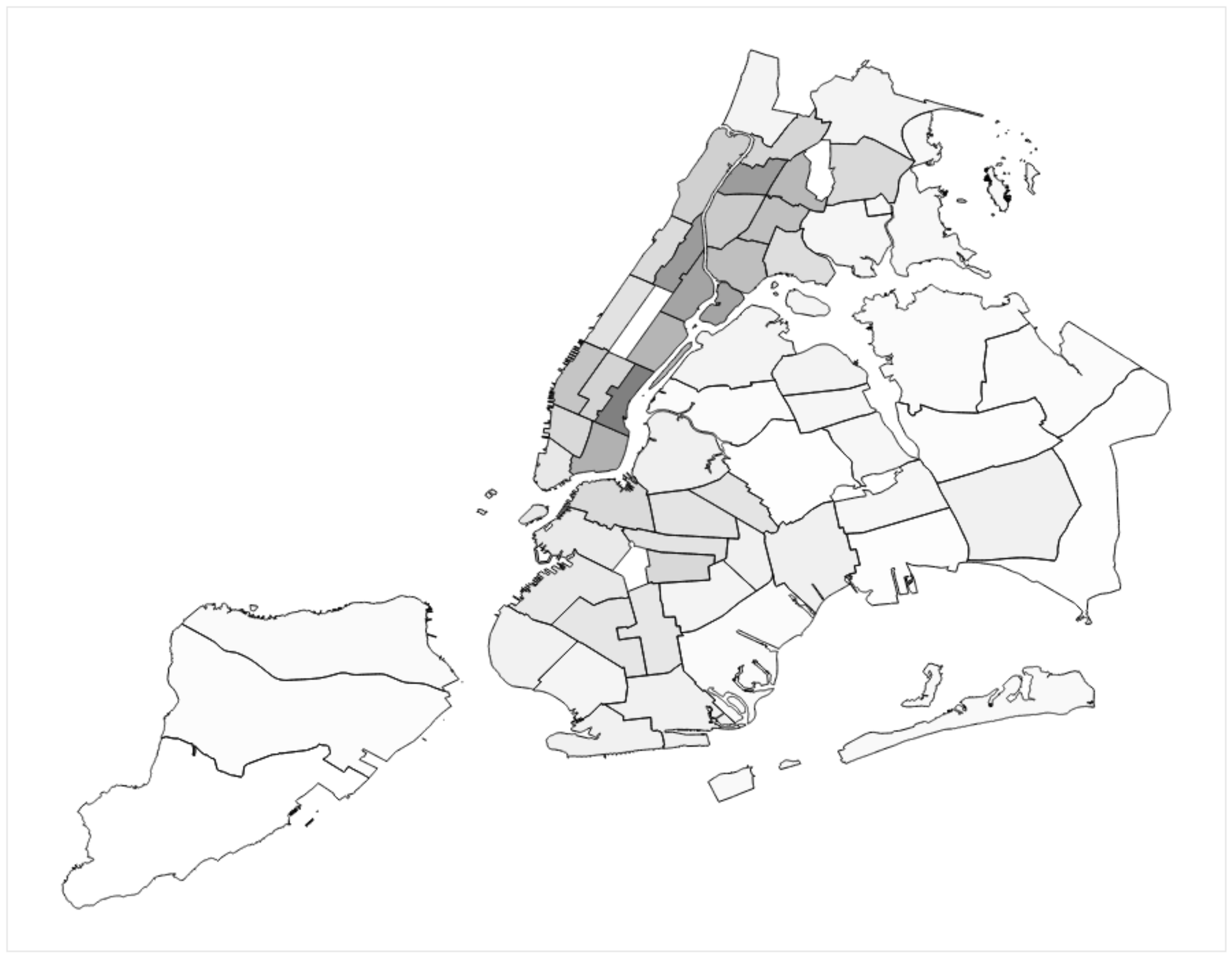}\label{fig:}}
 \vspace{-5pt}
 \caption{Top-2 auxiliary data sets ranked by the 2-stage SD for PM2.5 data set.}
 \label{fig:top-3_S} 
\end{figure}

{\bf Evaluation of auxiliary spatial data sets:}
Table~\ref{ranking} shows 
the top ten relevant auxiliary data sets as determined
by our model and 2-stage SD for the PM2.5 data set. 
These auxiliary data sets are arranged 
in descending order of the absolute values of 
the estimated regression coefficient ${\vect w}$, 
each of which is listed in the ``$w_s$'' columns of Table~\ref{ranking}.
By comparing the sorted list of the auxiliary data sets 
created by the proposed model 
with that yielded by 2-stage SD, 
we can confirm that 
the proposed model assigned relatively large regression coefficients to the auxiliary data sets with finer-grained partitions (i.e., Zip code and Taxi zone).

Figures~\ref{fig:top-3_P} and~\ref{fig:top-3_S} visualize 
the top two relevant auxiliary data sets 
as estimated by our model and 2-stage SD 
for the PM2.5 data set, respectively. 
Comparing these visualizations with that of the true target data in Figure~\ref{fig:PM2.5_true} 
shows that our model emphasized the most useful auxiliary data sets, i.e., those that are both strongly related with the target data and have fine granularities; 
2-stage SD evaluated the usefulness of auxiliary data sets 
only in terms of the strength of relationships
between the target data and the auxiliary data sets in the coarse-grained partition.

\begin{figure}[t]
 \centering
 \includegraphics[width=80mm]{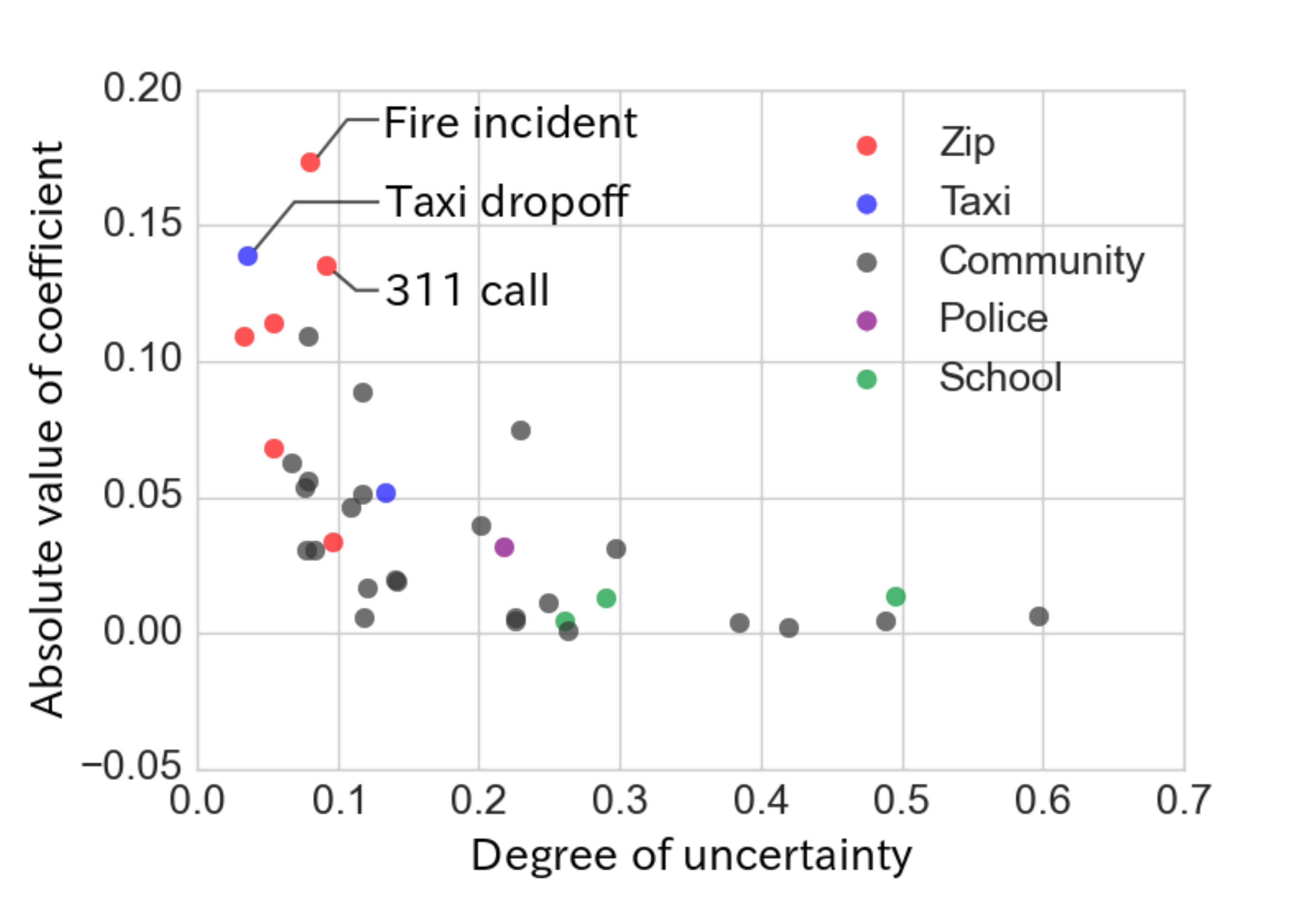}
 \vspace{-10pt}
 \caption{Relation between the coefficients and the uncertainties for PM2.5 data set.}
 \label{fig:relation}
\end{figure}

Figure~\ref{fig:relation} shows the relation between the regression coefficient and the uncertainty in the prediction of auxiliary data sets 
estimated by the proposed model for the PM2.5 data set.
In this figure, 
each auxiliary data set is depicted 
by a dot whose color indicates its partition. 
The horizontal axis shows the averages of the variances in the predicted values of each auxiliary data set; 
for the $s$th auxiliary data set, 
the average of variances was calculated by $(1/|{\mathcal P}^{\rm fine}|) 
\sum_{j \in {\mathcal P}^{\rm fine}} \Sigma_s^\ast (j,j)$, 
which is the degree of uncertainty in predicting the $s$th auxiliary data set; 
the vertical axis shows 
the absolute values of the estimated coefficients.
As shown, 
the absolute coefficient values
estimated by our model were likely to be higher for the auxiliary data sets that had lower degrees of uncertainty. 
These results indicate that our model can effectively learn the usefulness of each auxiliary data set by considering the uncertainty 
in the prediction of auxiliary data sets. 
Consequently, 
the proposed model can precisely refine the coarse-grained target data by effectively utilizing auxiliary data sets with various granularities. 

\section{Conclusion}
\label{sec:conclusion}
This paper has proposed a probabilistic model 
for refining coarse-grained spatial data 
by utilizing auxiliary spatial data sets 
with various granularities on the same region. 
Our model can effectively make use of auxiliary data sets 
with various granularities by hierarchically incorporating Gaussian processes.
Our model also has the advantage of allowing 
the inference of model parameters based on the exact marginal likelihood,
in which the variables of fine-grained target and auxiliary data are analytically 
integrated out.
Using multiple real-world spatial data sets in New York City, 
we confirmed that our model can predict the fine-grained target data 
more precisely compared with the baselines. 

Our future work is to consider shapes of regions as in the previous study~\cite{rathbun:spatial}:
The assumption of using the centroid of each region
allows for GP-based formulations 
and significantly simplifying computations involved;
meanwhile, it might worsen the fit of the GP 
to the exotic shaped regions (e.g., extremely elongated).
Another future work is to incorporate 
fully Bayesian treatment for model parameters. 
It can be expected to provide the better results.


\begin{thebibliography}{}

\bibitem[\protect\citeauthoryear{Barlacchi \bgroup et al\mbox.\egroup
  }{2015}]{barlacchi:multi}
Barlacchi, G.; Nadai, M.~D.; Larcher, R.; Casella, A.; Chitic, C.; Torrisi, G.;
  and et~al.
\newblock 2015.
\newblock A multi-source dataset of urban life in the city of {M}ilan and the
  province of {T}rentino.
\newblock {\em Scientific Data} 2.

\bibitem[\protect\citeauthoryear{Bogomolov \bgroup et al\mbox.\egroup
  }{2014}]{bogomolov:once}
Bogomolov, A.; Lepri, B.; Staiano, J.; Oliver, N.; Pianesi, F.; and Pentland,
  A.
\newblock 2014.
\newblock Once upon a crime: Towards crime prediction from demographics and
  mobile data.
\newblock In {\em ICMI},  427--434.

\bibitem[\protect\citeauthoryear{Boucher and Kyriakidis}{2006}]{boucher:super}
Boucher, A., and Kyriakidis, P.~C.
\newblock 2006.
\newblock Super-resolution land cover mapping with indicator geostatistics.
\newblock {\em Remote Sensing of Environment} 104:264--282.

\bibitem[\protect\citeauthoryear{Cannon}{2011}]{cannon:quantile}
Cannon, A.~J.
\newblock 2011.
\newblock Quantile regression neural networks: Implementation in {R} and
  application to precipitation downscaling.
\newblock {\em Computers \& Geosciences} 37(9):1277--1284.

\bibitem[\protect\citeauthoryear{Diodato \bgroup et al\mbox.\egroup
  }{2010}]{diodato:multiscale}
Diodato, N.; Bellocchi, G.; Bertolin, C.; and Camuffo, D.
\newblock 2010.
\newblock Multiscale regression model to infer historical temperatures in a
  central mediterranean sub-regional area.
\newblock {\em Climate of the Past Discussions} 6:2625--2649.

\bibitem[\protect\citeauthoryear{Dong \bgroup et al\mbox.\egroup
  }{2014}]{dong:learning}
Dong, C.; Loy, C.~C.; He, K.; and Tang, X.
\newblock 2014.
\newblock Learning a deep convolutional network for image super-resolution.
\newblock In {\em ECCV},  184--199.

\bibitem[\protect\citeauthoryear{Flaxman, Wang, and Smola}{2015}]{flaxman:who}
Flaxman, S.~R.; Wang, Y.~X.; and Smola, A.~J.
\newblock 2015.
\newblock Who supported {O}bama in 2012?: Ecological inference through
  distribution regression.
\newblock In {\em KDD},  289--298.

\bibitem[\protect\citeauthoryear{Ghosh}{2010}]{ghosh:SVM}
Ghosh, S.
\newblock 2010.
\newblock {SVM}-{PGSL} coupled approach for statistical downscaling to predict
  rainfall from {GCM} output.
\newblock {\em Journal of Geophysical Research: Atmospheres} 115(D22).

\bibitem[\protect\citeauthoryear{Goldstein and Dyson}{2013}]{goldstein:beyond}
Goldstein, B., and Dyson, L.
\newblock 2013.
\newblock Beyond transparency: Open data and the future of civic innovation.
\newblock {\em Code for America Press}.

\bibitem[\protect\citeauthoryear{Goovaerts}{2010}]{goovaerts:combining}
Goovaerts, P.
\newblock 2010.
\newblock Combining areal and point data in geostatistical interpolation:
  Applications to soil science and medical geography.
\newblock {\em Mathematical Geosciences} 42(5):535--554.

\bibitem[\protect\citeauthoryear{Hessami \bgroup et al\mbox.\egroup
  }{2008}]{hessami:automated}
Hessami, M.; Gachon, P.; Ouarda, T.~B.; and St-Hilair, A.
\newblock 2008.
\newblock Automated regression-based statistical downscaling tool.
\newblock {\em Environmental Modeling \& Software} 23(6):813--834.

\bibitem[\protect\citeauthoryear{Howitt and Reynaud}{2003}]{howitt:spatial}
Howitt, R., and Reynaud, A.
\newblock 2003.
\newblock Spatial disaggregation of agricultural production data using maximum
  entropy.
\newblock {\em European Review of Agricultural Economics} 30(2):359--387.

\bibitem[\protect\citeauthoryear{Jerrett \bgroup et al\mbox.\egroup
  }{2013}]{jerrett:spatial}
Jerrett, M.; Burnett, R.~T.; Beckerman, B.~S.; Turner, M.~C.; Krewski, D.;
  Thurston, G.; and et~al.
\newblock 2013.
\newblock Spatial analysis of air pollution and mortality in {C}alifornia.
\newblock {\em American Journal of Respiratory and Critical Care Medicine}
  188(5):593--599.

\bibitem[\protect\citeauthoryear{Keil \bgroup et al\mbox.\egroup
  }{2013}]{keil:downscaling}
Keil, P.; Belmaker, J.; Wilson, A.~M.; Unitt, P.; and Jetz, W.
\newblock 2013.
\newblock Downscaling of species distribution models: a hierarchical approach.
\newblock {\em Methods in Ecology and Evolution} 4(1):82--94.

\bibitem[\protect\citeauthoryear{Kyriakidis}{2004}]{kyriakidis:geostatistical}
Kyriakidis, P.~C.
\newblock 2004.
\newblock A geostatistical framework for area-to-point spatial interpolation.
\newblock {\em Geographical Analysis} 36(3):259--289.

\bibitem[\protect\citeauthoryear{Law \bgroup et al\mbox.\egroup
  }{2018}]{Law:variational}
Law, H. C.~L.; Sejdinovic, D.; Cameron, E.; Lucas, T. C.~D.; Flaxman, S.;
  Battle, K.; and Fukumizu, K.
\newblock 2018.
\newblock Variational learning on aggregate outputs with {G}aussian processes.
\newblock In {\em NeurIPS}, 6084--6094.

\bibitem[\protect\citeauthoryear{Liu and Nocedal}{1989}]{liu:on}
Liu, D.~C., and Nocedal, J.
\newblock 1989.
\newblock On the limited memory {BFGS} method for large scale optimization.
\newblock {\em Mathematical programming} 45(1--3):503--528.

\bibitem[\protect\citeauthoryear{Miller \bgroup et al\mbox.\egroup
  }{2015}]{miller:impact}
Miller, B.~A.; Koszinski, S.; Wehrhan, M.; and Sommer, M.
\newblock 2015.
\newblock Impact of multi-scale predictor selection for modeling soil
  properties.
\newblock {\em Geoderma} 239:97--106.

\bibitem[\protect\citeauthoryear{Misra, Sarkar, and
  Mitra}{2017}]{misra:statistical}
Misra, S.; Sarkar, S.; and Mitra, P.
\newblock 2017.
\newblock Statistical downscaling of precipitation using long short-term memory
  recurrent neural networks.
\newblock {\em Theoretical and Applied Climatology} 134.

\bibitem[\protect\citeauthoryear{Murakami and Tsutsumi}{2011}]{murakami:new}
Murakami, D., and Tsutsumi, M.
\newblock 2011.
\newblock A new areal interpolation technique based on spatial econometrics.
\newblock {\em Procedia-Social and Behavioral Sciences} 21:230--239.

\bibitem[\protect\citeauthoryear{Park}{2013}]{park:spatial}
Park, N.~W.
\newblock 2013.
\newblock Spatial downscaling of {TRMM} precipitation using geostatistics and
  fine scale environmental variables.
\newblock {\em Advances in Meteorology} 2013.

\bibitem[\protect\citeauthoryear{Rasmussen and Williams}{2006}]{carl:gaussian}
Rasmussen, C.~E., and Williams, C. K.~I.
\newblock 2006.
\newblock {\em Gaussian processes for machine learning}.
\newblock MIT Press.

\bibitem[\protect\citeauthoryear{Rathbun}{1998}]{rathbun:spatial}
Rathbun, S.~L.
\newblock 1998.
\newblock Spatial modelling in irregularly shaped regions: Kriging estuaries.
\newblock {\em Environmetrics} 9:109--129.

\bibitem[\protect\citeauthoryear{Rupasinghaa and
  Goetz}{2007}]{rupasinghaa:social}
Rupasinghaa, A., and Goetz, S.~J.
\newblock 2007.
\newblock Social and political forces as determinants of poverty: A spatial
  analysis.
\newblock {\em The Journal of Socio-Economics} 36(4):650--671.

\bibitem[\protect\citeauthoryear{Shadbolt \bgroup et al\mbox.\egroup
  }{2012}]{shadbolt:linked}
Shadbolt, N.; O'Hara, K.; Berners-Lee, T.; Gibbins, N.; Glaser, H.; Wendy, H.;
  and Schraefel, M.~C.
\newblock 2012.
\newblock Linked open government data: Lessons from data.gov.uk.
\newblock {\em IEEE Intelligent Systems} 27(3):16--24.

\bibitem[\protect\citeauthoryear{Smith and Capra}{2016}]{smith:beyond}
Smith, C.~C., and Capra, L.
\newblock 2016.
\newblock Beyond the baseline: Establishing the value in mobile phone based
  poverty estimates.
\newblock In {\em WWW},  425--434.

\bibitem[\protect\citeauthoryear{Smith, Mashhadi, and
  Capra}{2014}]{Smith:poverty}
Smith, C.~C.; Mashhadi, A.; and Capra, L.
\newblock 2014.
\newblock Poverty on the cheap: Estimating poverty maps using aggregated mobile
  communication networks.
\newblock In {\em CHI},  511--520.

\bibitem[\protect\citeauthoryear{Stein}{1999}]{stein:interpolation}
Stein, M.~L.
\newblock 1999.
\newblock {\em Interpolation of spatial data: Some theory for kriging}.
\newblock Springer.

\bibitem[\protect\citeauthoryear{Sturrock \bgroup et al\mbox.\egroup
  }{2014}]{sturrock:fine}
Sturrock, H. J.~W.; Cohen, J.~M.; Keil, P.; Tatem, A.~J.; Menach, A.~L.;
  Ntshalintshali, N.~E.; Hsiang, M.~S.; and Gosling, R.~D.
\newblock 2014.
\newblock Fine-scale malaria risk mapping from routine aggregated case data.
\newblock {\em Malaria Journal} 13:421.

\bibitem[\protect\citeauthoryear{Taylor, Andrade-Pacheco, and
  Sturrock}{2018}]{taylor:continuous}
Taylor, B.~M.; Andrade-Pacheco, R.; and Sturrock, H. J.~W.
\newblock 2018.
\newblock Continuous inference for aggregated point process data.
\newblock {\em Journal of the Royal Statistical Society: Series A (Statistics
  in Society)} 181(4):1125--1150.

\bibitem[\protect\citeauthoryear{Vandal \bgroup et al\mbox.\egroup
  }{2017}]{vandal:deepsd}
Vandal, T.; Kodra, E.; Ganguly, S.; Michaelis, A.; Nemani, R.; and Ganguly,
  A.~R.
\newblock 2017.
\newblock Deep{SD}: Generating high resolution climate change projections
  through single image super-resolution.
\newblock In {\em KDD},  1663--1672.

\bibitem[\protect\citeauthoryear{Vandal \bgroup et al\mbox.\egroup
  }{2018}]{vandal:generating}
Vandal, T.; Kodra, E.; Ganguly, S.; Michaelis, A.; Nemani, R.; and Ganguly,
  A.~R.
\newblock 2018.
\newblock Generating high resolution climate change projections through single
  image super-resolution: An abridged version.
\newblock In {\em IJCAI},  5389--5393.

\bibitem[\protect\citeauthoryear{Wang \bgroup et al\mbox.\egroup
  }{2016}]{wang:crime}
Wang, H.; Kifer, D.; Graif, C.; and Li, Z.
\newblock 2016.
\newblock Crime rate inference with big data.
\newblock In {\em KDD},  635--644.

\bibitem[\protect\citeauthoryear{Wilby \bgroup et al\mbox.\egroup
  }{2004}]{wilby:guidelines}
Wilby, R.~L.; Zorita, S.~P.; Timbal, E.; Whetton, B.; and Mearns, L.~O.
\newblock 2004.
\newblock Guidelines for Use of Climate Scenarios Developed from
  Statistical Downscaling Methods.
\newblock {\em Supporting material of the IPCC}.

\bibitem[\protect\citeauthoryear{Wilson and Wakefield}{2018}]{wilson:pointless}
Wilson, K., and Wakefield, J.
\newblock 2018.
\newblock Pointless spatial modeling.
\newblock {\em Biostatistics}.

\bibitem[\protect\citeauthoryear{Wotling \bgroup et al\mbox.\egroup
  }{2000}]{wotling:regionalization}
Wotling, G.; Bouvier, C.; Danloux, J.; and Fritsch, J.~M.
\newblock 2000.
\newblock Regionalization of extreme precipitation distribution using the
  principal components of the topographical environment.
\newblock {\em Journal of Hydrology} 233:86--101.

\bibitem[\protect\citeauthoryear{Xavier \bgroup et al\mbox.\egroup
  }{2016}]{xavier:disaggregating}
Xavier, A.; Freitas, M. B.~C.; Ros$\rm{j\acute{a}}$rio, M. D.~S.; and Fragoso,
  R.
\newblock 2016.
\newblock Disaggregating statistical data at the field level: An entropy
  approach.
\newblock {\em Spatial Statistics} 23:91--103.

\bibitem[\protect\citeauthoryear{Xu \bgroup et al\mbox.\egroup
  }{2018}]{xu:muscat}
Xu, J.; Liu, X.; Wilson, T.; Tan, P.~N.; Hatami, P.; and Luo, L.
\newblock 2018.
\newblock Muscat: Multi-scale spatio-temporal learning with application to
  climate modeling.
\newblock In {\em IJCAI},  2912--2918.

\bibitem[\protect\citeauthoryear{Xu}{2017}]{xu:multi}
Xu, J.
\newblock 2017.
\newblock Multi-task learning and its application to geospatio-temporal data.
\newblock {\em ProQuest Dissertations Publishing}.

\bibitem[\protect\citeauthoryear{Yuan, Zheng, and Xie}{2012}]{yuan:discovering}
Yuan, J.; Zheng, Y.; and Xie, X.
\newblock 2012.
\newblock Discovering regions of different functions in a city using human
  mobility and {poi}s.
\newblock In {\em KDD},  186--194.

\bibitem[\protect\citeauthoryear{Zheng \bgroup et al\mbox.\egroup
  }{2015}]{zheng:forecasting}
Zheng, Y.; Yi, X.; Li, M.; Li, R.; Shan, Z.; Chang, E.; and Li, T.
\newblock 2015.
\newblock Forecasting fine-grained air quality based on big data.
\newblock In {\em KDD},  2267--2276.

\bibitem[\protect\citeauthoryear{Zheng, Liu, and Hsieh}{2013}]{zheng:U-Air}
Zheng, Y.; Liu, F.; and Hsieh, H.~P.
\newblock 2013.
\newblock U-air: When urban air quality inference meets big data.
\newblock In {\em KDD},  1436--1444.

\bibitem[\protect\citeauthoryear{Zorita and Storch}{1999}]{zorita:analog}
Zorita, E., and Storch, H. V.
\newblock 1999.
\newblock The analog method as a simple statistical downscaling technique:
  Comparison with more complicated methods.
\newblock {\em Journal of Climate} 12:2474--2489.

\end{thebibliography}

\appendix
\section{Derivatives of model parameters}
\label{sec:derivative}
The log-marginal likelihood of ${\vect a}$ is given by
\vspace{-0.15\baselineskip}
\begin{flalign}
 \log p({\vect a} \mid {\vect w}, \alpha, \gamma, \sigma) 
 &= -\frac{1}{2}\T{({\vect a} - {\bf H}\bar{{\bf F}}^\ast{\vect w})} {\vect \Lambda}^{-1}
  ({\vect a} - {\bf H}\bar{{\bf F}}^\ast{\vect w}) \nonumber \\
 &-\frac{1}{2} \log\left( \det({\vect \Lambda})\right) 
  - \frac{|{\mathcal P}^{\rm coar}|}{2} \log 2\pi.
 \label{eq:objective}
\end{flalign}
We describe the first derivatives of~(\ref{eq:objective}) 
with respect to $w_s$, $\alpha$, $\gamma$, $\sigma$, 
which is required for estimating the parameter 
based on the BFGS method.
The derivative of (\ref{eq:objective}) with respect to $w_s$
is given by
\vspace{-0.15\baselineskip}
\begin{flalign}
 &\frac{\partial}{\partial w_s} \log p({\vect a} \mid {\vect w}, \alpha, \gamma, \sigma) \nonumber \\
 &= \frac{\partial ({\vect a}-{\bf H}\bar{{\bf F}}^\ast{\vect w})}{\partial w_s}{\vect p}
 + \frac{1}{2}{\rm tr} 
 \left(({\vect p}\T{{\vect p}} - {\vect \Lambda}^{-1}) \frac{\partial {\vect \Lambda}}
  {\partial w_s}\right), 
\end{flalign}
where ${\vect p} = {\vect \Lambda}^{-1} ({\vect a}-{\bf H}\bar{{\bf F}}^\ast{\vect w})$ 
and $\partial {\vect \Lambda} / \partial w_s$ is a matrix of elementwise derivatives. 
The derivative of the element $\Lambda(i,i^\prime)$~(\ref{eq:lambda}) is obtained by
\vspace{-0.15\baselineskip}
\begin{equation}
 \frac{\partial \Lambda(i,i^\prime)}{\partial w_s} 
  = \frac{1}{|{\mathcal P}^{\rm fine}_i||{\mathcal P}^{\rm fine}_{i^\prime}|}
  \sum_{j \in {\mathcal P}^{\rm fine}_i} 
  \sum_{j^\prime \in {\mathcal P}^{\rm fine}_{i^\prime}}
  2 w_s \Sigma_s^\ast(j,j^\prime).
\end{equation}
Denoting $\theta \in \{\alpha, \gamma, \sigma\}$, 
the derivative of (\ref{eq:objective}) with respect to $\theta$ is given by
\vspace{-0.15\baselineskip}
\begin{equation}
 \frac{\partial}{\partial \theta} 
  \log p({\vect a} \mid {\vect w}, \alpha, \gamma, \sigma)
 = \frac{1}{2}{\rm tr}
 \left(({\vect p}\T{{\vect p}} - {\vect \Lambda}^{-1}) 
  \frac{\partial {\vect \Lambda}}{\partial \theta}\right). 
\end{equation}
The matrix of elementwise derivatives 
$\partial {\vect \Lambda} / \partial \theta$ is trivial. 
The derivative of the element $\Lambda(i,i^\prime)$~(\ref{eq:lambda}) 
with respect to each hyperparameter 
is as follows: 
\vspace{-0.15\baselineskip}
\begin{flalign}
 &\frac{\partial \Lambda(i,i^\prime)}{\partial \alpha} \nonumber \\
 &= \frac{1}{|{\mathcal P}^{\rm fine}_i||{\mathcal P}^{\rm fine}_{i^\prime}|} 
  \sum_{j \in {\mathcal P}^{\rm fine}_i} 
 \sum_{j^\prime \in {\mathcal P}^{\rm fine}_{i^\prime}}
 2\alpha \exp\left(-\frac{1}{2\gamma^2} \|{\vect x}_j - {\vect x}_{j^\prime}\|^2\right),
\end{flalign}
\vspace{-0.15\baselineskip}
\begin{flalign}
 \frac{\partial \Lambda(i,i^\prime)}{\partial \gamma} 
 &= \frac{1}{|{\mathcal P}^{\rm fine}_i||{\mathcal P}^{\rm fine}_{i^\prime}|}
  \sum_{j \in {\mathcal P}^{\rm fine}_i} 
  \sum_{j^\prime \in {\mathcal P}^{\rm fine}_{i^\prime}} 
  \alpha^2 \left(\frac{1}{\gamma^3} \|{\vect x}_j - {\vect x}_{j^\prime}\|^2\right) \nonumber \\
 &\times \exp\left(-\frac{1}{2\gamma^2} \|{\vect x}_j - {\vect x}_{j^\prime}\|^2\right),
\end{flalign}
\vspace{-0.15\baselineskip}
\begin{flalign}
 \frac{\partial \Lambda(i,i^\prime)}{\partial \sigma} = 2\sigma \delta_{i,i^\prime}.
\end{flalign}

\section{Description of real-world spatial data sets}
\label{sec:discription}
We used the real-world spatial data sets from NYC Open Data~\footnote{https://opendata.cityofnewyork.us}.
for evaluating the proposed model. 
The data sets were collected and released for improving the urban environment in New York City, 
and contain a variety of categories 
such as social indicators, land use, air quality and taxi traffic. 
Details of the data sets are listed in Table~\ref{Spatial Data}. 
There are multiple data sets in each category, with the total number of data sets being 44. 
Each data set is associated with one of six geographical partitions, i.e.,
school district, UHF42, community district, police precinct, zip code and taxi zone.
These partitions have various spatial granularities; 
the number of regions in each partition is shown in Table~\ref{Spatial Data}. 
These data sets are gathered once a year using the time ranges shown in Table~\ref{Spatial Data}; 
the values of data are divided by the number of observation times. 
When the values of data are extensive quantities (i.e., proportional to the scale of areas, e.g., population), 
the values are divided by the areas of respective regions; 
the resulting values are intensive quantities (i.e., independent of area scale, e.g., population density). 

In our experiments, 
we try to refine the poverty rate data set in the social indicator category 
and the five air pollution data sets in the air quality category. 
The poverty rate data set contains the values of poverty rates associated with each region in the community district partition
as visualized in Figure~\ref{fig:intro_fig_comm}. 
The air pollution data sets 
contain the average concentrations of pollutants 
(i.e., PM2.5, ozone, formaldehyde, benzene, elemental carbon) 
associated with each region in the UHF42 partition. 
In order to evaluate the performance in refining coarse-grained data, we used the data that were aggregated 
into a coarser-grained partition, i.e., borough partition, via spatial averaging, 
where the borough partition has five regions as illustrated in Figure~\ref{fig:intro_fig_boro}.
The experimental setting is as follows: 
1) Given the poverty rate data set with borough partition ($|{\mathcal P}^{\rm coar}| = 5$), 
we would like to refine the data into the community district partition ($|{\mathcal P}^{\rm fine}| = 59$), 
and 2) given each air pollution data set with the borough partition ($|{\mathcal P}^{\rm coar}| = 5$), 
we aim to refine the data into the UHF42 partition ($|{\mathcal P}^{\rm fine}| = 42$). 
In the setting for the poverty rate data set, we used all data sets other than the target data as auxiliary data sets, 
so the number of auxiliary data sets $|{\mathcal S}|$ was 43. 
In the setting for the air pollution data sets, we used all data sets not contained in the air quality category, so $|{\mathcal S}|$ was 36. 
\begin{table*}[t]
\caption{Spatial data sets.}
\vspace{-10pt}
\begin{center}
\small
\begin{tabular}{l r l r c l} \hline
Category/Name &\#data sets &Partition &\#regions &Time range &\multicolumn{1}{c}{Description} \\ \hline
Education &3 &School district &32 &2010 &Class size, ratio of \#pupils to \#teachers, SAT score \\
Air quality &8  &UHF42 &42 &2009--2010 &Average concentration of pollutants \\
Social indicator &13 &Community district &59 &2009--2013 &Poverty rate, population, mean commute time, etc.  \\
Land use &11 &Community district &59 &2009--2013 &Area percentage for commercial office, parking, etc.  \\
Crime &1 &Police precinct &77 &2010--2016 &Number of crimes \\
Incident &2 &Zip code &186 &2010--2016 &\#311 calls, \#fire incidents \\
Telecommunication &2 &Zip code &186 &2016 &\#public telephones, \#free Wi-Fi hotspots \\
Consumption &2 &Zip code &186 &2010--2014 &GHG emission, natural gas consumption\\
Taxi traffic &2 &Taxi zone &249 &2014--2016 &\#taxi pick-up and drop-off events \\
\hline
\end{tabular}
\label{Spatial Data}
\end{center}
\vspace{-10pt}
\end{table*}

\section{Baselines description}
\label{sec:baselines}
For GPR, we predict the fine-grained target data ${\vect z}$
based only on the coarse-grained target data ${\vect a}$. 
For LR-based method and 2-stage SD, 
given the coarse-grained target data ${\vect a}$ 
and the predictive values of all auxiliary data sets $\bar{{\bf F}}^\ast$, 
we predict the fine-grained target data ${\vect z}$. 
Details of these baselines are given below.

{\bf Gaussian process regression (GPR)}: 
We compared our proposed model with a simple spatial interpolation (i.e., GPR) 
of the coarse-grained spatial data ${\vect a}$. 
This baseline assumes that the target data are explained 
by only the spatial correlation. 
Given ${\vect a}$ and the set of centroids of 
the coarse-grained partition ${\mathcal P}^{\rm coar}$, 
we predicted the fine-grained target data ${\vect z}$ 
by using the predictive distribution. 
Note that this baseline does not use the auxiliary spatial data sets. 

{\bf Linear regression-based method (LR-based method)}: 
We used a linear regression-based method that has been applied in various studies~\cite{bogomolov:once,Smith:poverty}.
The linear regression model is used for estimating the relationships between the coarse-grained target data and the auxiliary data sets. 
The procedure in the training phase is as follows: 
1) aggregate all auxiliary data sets into the coarse-grained partition of target data via spatial averaging; 
2) estimate the regression coefficients ${\vect w}$ of the respective auxiliary data sets by using the coarse-grained target data 
and the auxiliary data sets aggregated via spatial averaging. 
In the prediction phase, 
generate unknown values ${\vect z}$ for the target fine-grained partition by applying the estimated relationships to the predictive values of 
auxiliary data sets $\bar{{\bf F}}^\ast$ as follows:
${\vect z} = \bar{{\bf F}}^\ast \hat{{\vect w}}$, where $\hat{{\vect w}}$ is the estimated regression coefficient. 

{\bf Two-stage statistical downscaling method (2-stage SD)}: 
We used the statistical downscaling method proposed in~\cite{park:spatial}. 
This method assumes that coarse-grained target data ${\vect a}$ can be decomposed into linear regression terms and residual terms. 
The downscaling procedure is divided into two stages. 
In the first stage, 
we obtain the regression coefficients ${\vect w}$ in a manner similar to the training phase of the LR-based method. 
In the second stage, 
given the estimated coefficient $\hat{{\vect w}}$, 
the fine-grained target data ${\vect z}$ are estimated to be those that satisfy the following relation:
\vspace{-0.15\baselineskip}
\begin{flalign}
 a_i 
 &= \underbrace{
 \hat{w}_0 + \sum_{s \in {\mathcal S}} \hat{w}_s 
 \left[
 \frac{1}{|{\mathcal P}^{\rm fine}_i|} \sum_{j \in {\mathcal P}^{\rm fine}_i} f_s({\vect x}_j)
 \right]}_{\text{linear regression term}}
 + \underbrace{
 R_i^{\rm coar}}_{\text{residual term}} \nonumber \\
 &= \frac{1}{|{\mathcal P}^{\rm fine}_i|} \sum_{j \in {\mathcal P}^{\rm fine}_i}
 \left[
 \hat{w}_0 + \sum_{s \in {\mathcal S}} \hat{w}_s f_s({\vect x}_j) + R_j^{\rm fine}
 \right] \nonumber \\
 &= \frac{1}{|{\mathcal P}^{\rm fine}_i|} \sum_{j \in {\mathcal P}^{\rm fine}_i} z_j. 
 \label{eq:SD}
\end{flalign}
This relation expresses the spatial aggregation constraint, i.e., the assumption that value $a_i$ associated with coarse-grained region $i$ 
is the linear average of the constituent values in the fine-grained partition. 
Here, $R_i^{\rm coar}$ and $R_j^{\rm fine}$ are the residuals in the coarse-grained and fine-grained partitions, respectively. 
To obtain the fine-grained target data ${\vect z}$, the residual value $R_j^{\rm fine}$ in the fine-grained partition must be determined. 
Since the linear regression terms have already been fixed in the first stage, $R_i^{\rm coar}$ is obtained from (\ref{eq:SD}); 
the residuals in the fine-grained partition are predicted by applying the spatial interpolation method, 
i.e., simple kriging~\cite{kyriakidis:geostatistical}, 
to the residuals $R_i^{\rm coar}$ in the coarse-grained partition. 

 \end{document}